%% file: main.tex
\documentclass{article}

\PassOptionsToPackage{numbers, compress}{natbib}
\usepackage[preprint]{neurips_2024}

\newcommand{\LLMagent}{\texttt{StaffPro}\xspace}

\usepackage{packages}

\newcommand{\code}[1]{\texttt{#1}}
\newcommand{\boldcode}[1]{{\ttfamily \fontseries{b} \selectfont #1}}
\newcommand{\doubleapex}[1]{``#1''}
\DeclareMathOperator*{\argmax}{arg\,max}

\theoremstyle{definition} % possible values: definition, plain, remark
\newtheorem{definition}{Definition}
\newtheorem*{definition*}{Definition}

\title{\texttt{StaffPro:} an LLM Agent for Joint Staffing and Profiling}

\author{%
  Alessio Maritan\\
  University of Padova\\
  \texttt{alessio.maritan@phd.unipd.it} \\
}

\begin{document}

\maketitle

\begin{abstract}
Large language model (LLM) agents integrate pre-trained LLMs with modular algorithmic components and have shown remarkable reasoning and decision-making abilities. In this work, we investigate their use for two tightly intertwined challenges in workforce management: \textit{staffing}, i.e., the assignment and scheduling of tasks to workers, which may require team formation; and \textit{profiling}, i.e., the continuous estimation of workers' skills, preferences, and other latent attributes from unstructured data. We cast these problems in a formal mathematical framework that links scheduling decisions to latent feature estimation, and we introduce \LLMagent, an LLM agent that addresses staffing and profiling jointly.
Differently from existing staffing solutions, \LLMagent allows expressing optimization objectives using natural language, accepts textual task descriptions and provides high flexibility.
\LLMagent interacts directly with humans by establishing a continuous human-agent feedback loop, ensuring natural and intuitive use.
By analyzing human feedback, our agent continuously estimates the latent features of workers, realizing life-long worker profiling and ensuring optimal staffing performance over time.
A consulting firm simulation example demonstrates that \LLMagent successfully estimates workers' attributes and generates high quality schedules.
With its innovative design, \LLMagent offers a robust, interpretable, and human-centric solution for automated personnel management.
\end{abstract}

\section{Introduction}
One of the newest and most promising paradigms in the field of artificial intelligence is that of large language model (LLM) agents, i.e. autonomous agents arising from the combination of LLMs with other software tools that extend their capabilities, allowing them to interact with an environment to achieve given objectives.
Showing impressive reasoning, generation and decision-making capabilities, LLM agents open up new possibilities to address unsolved and highly challenging problems in automation. One of the most relevant challenges in this area is the control and performance optimization of complex, multi-agent, time-varying systems with {humans-in-the-loop}, such as companies and large-scale human organizations.

Motivated by this, in this work we explore the potential of LLM agents to assist in the management of personnel in companies and organizations, focusing on two relevant problems: \textit{staffing} and \textit{profiling}.
Staffing consists in task assignment and scheduling, involving team formation in case tasks require multiple workers. Informally, given a list of tasks and a set of human workers, the staffing problem lies within finding the best way of distributing tasks among workers over a time horizon to pursue some given objectives.
Profiling consists in creating and continuously updating a structured description of workers that includes technical and interpersonal skills, personal preferences and any other information relevant to their functionality in the system. We argue that LLM agents are very promising candidates to address these problems: with their semantic understanding, deductive reasoning and built-in knowledge, properly designed LLM agents can extensively process the large body of data generated in work environments, automate repetitive analyses that would otherwise require humans, generate optimal schedules and provide updated estimates of workers' characteristics.
Below we highlight the importance of solving staffing and profiling in an automated way and describe the intrinsic relationship between them. Then, we introduce \LLMagent, our proposed LLM agent which addresses these problems jointly, and we summarize its benefits and innovative design compared to existing literature.
%\ndf{OPTIONAL: after putting staffing and profiling in a more joint fashion, we could say: we formulate this problem formally as an exploration-exploitation paradigm, in which the purpose is to jointly learn the features of the workers from experience, feedback and interactions (exploration) while optimizing their utilization (exploitation). We could also add: we argue that while a reinforcement learning strategy for this problem would be possible, its implementation would be infeasible/impractical and ultimately useless. Why? Because (i) RL is mostly trained in simulations and (ii) RL requires a lot of data samples and (iii) workers and organizations change structure and personnel often, time-varying, dramatic distribution shifts... and also the contraints...}

\textbf{Staffing.}
Staffing is a frequent necessity in work environments that adopt task-driven workflows, where employees are assigned to tasks or temporary project teams based on need. Properly solving staffing problems can be difficult due to conflicting objectives, constraints and social factors, especially in large companies. For example, in consulting companies or outsourcing agencies where tasks are commissioned by external clients, one may want to assemble the most suitable team of consultants for each task based on their skills, preferences and reciprocal compatibility. In healthcare, tasks can be patients requiring medical treatment, and one can desire to assign to each patient the doctor whose specializations and expertise best fit the task, while also minimizing the overall waiting time of patients.
Since staffing is a complex combinatorial and constrained problem, it is desirable to perform it in an automated way, which can also ensure systematic application of decision criteria and remove subjectivity in decision making. However, the currently available options for automated staffing only support a small set of predefined optimization objectives, that must necessarily be expressed in analytical form. In all other cases, tasks must be staffed and scheduled manually, which can be time-consuming and lead to suboptimal solutions.
To address this problem, in this work we leverage an LLM to automate staffing for arbitrary optimization objectives, that can be partially or entirely expressed using natural language. This removes the need to engineer complex utility functions and enables choosing objectives that would be difficult or impossible to encode using only mathematical tools.

\textbf{Profiling.}
Profiling workers involves estimating some of their characteristics that cannot be accessed directly, such as their proficiency levels in technical skills, personal preferences and the prevalence of certain personality traits.
Data of this type is valuable to companies, who use it for strategic planning, internal knowledge assessment and general workforce management.
For example, estimating the skills of workers results in a skill matrix, a visualization tool commonly used by organizations to check the competences of all employees at a glance. The matrix provides an overview of the talent pool of the organization, helping in identifying skill gaps and hiring needs, planning staff training and deciding about career advancements.
In practice, worker profiling is done by collecting and analyzing information related to the workers' characteristics to be estimated, such as reports, feedback from various people and empirical evidence. The clues extracted from different information sources must then be compared to formulate beliefs about workers and constantly update them.
This process can be time-consuming and subject to cognitive biases if done manually, thus motivating fully automated worker profiling.

\textbf{Joint staffing and profiling.}
Staffing and worker profiling naturally benefit each other in a virtuous loop: knowledge about workers can improve the quality and effectiveness of staffing solutions, and feedback about task assignments can provide additional information about workers.
Even more so, if staffing criteria concern some attributes of workers whose real value is unknown and must be estimated, then profiling workers becomes a requirement. In any case, profiling widens the range of objectives that can be optimized in staffing, allowing to consider individual characteristics of workers such as their personality and preferences, that are known to be critical factors for job satisfaction, performance and teamwork efficacy \cite{ivancevich1990organizational}. For example, simply assigning tasks to the most qualified employees and neglecting personal preferences can be suboptimal, as workers who like their job tend to be more productive \cite{isen1991positive}. As staffing and worker profiling are so tightly interconnected, it is intuitive to address them together.
In our automated framework, staffing and worker profiling are synergistic and rely on each other for continuous improvement in a closed-loop fashion. We consider staffing problems in which some of the objectives to be optimized are implicitly formulated in terms of the true attributes of workers, which are unknown. By estimating these attributes, worker profiling promotes the correct evaluation of criteria and objectives, and therefore the selection of schedules that faithfully align with the intentions of human supervisors. Vice versa, the reports and feedback obtained as by-products of staffing are used to profile workers, enabling our LLM agent to learn, stay updated and improve over time.

%Existing automated approaches to solve staffing problems \cite{bahroun2024multi} are based on restrictive assumptions, require rigid models of tasks and workers, and are limited to very specific optimization objectives, thus lacking descriptive power and flexibility.
\textbf{Novelty.}
To the best of our knowledge, all existing automated approaches to solving staffing problems are based on classical optimization, and inherit its disadvantages: they impose restrictive assumptions, require rigid models of tasks and workers, and are limited to very specific optimization objectives \cite{bahroun2024multi}, thus lacking descriptive power and flexibility. The only partial exception is \cite{mikhridinova2024using}, which describes a single use case where an LLM is prompted to analyze the compatibility of two predefined teams for a specific project. Their study admittedly considers a scenario with insufficient complexity and provides unusable decision support.
In comparison, our LLM agent \LLMagent is a game changer for staffing problems. Recognizing the current limitations of LLMs, we do not simply prompt an LLM to solve the whole problem, but rather we use it as a building block of a larger system, which contains multiple tools and implements structured pipelines. Our solution combines the reliability and interpretability of standard optimization algorithms with the reasoning and generation capabilities of LLMs, taking the best of both worlds. 

\textbf{\LLMagent in a nutshell.}
We now briefly mention the main advantages of our approach, leaving a more detailed description for later.
With \LLMagent, you can specify an arbitrary set of objectives to be maximized by the staffing solution to be generated. These optimization objectives can be conveniently expressed using natural language, dispensing with handcrafted ad-hoc cost functions and broadening the range of optimization objectives that can be pursued. 
Additionally, the need for data preparation is eliminated: by leveraging the LLM, our agent can quickly process large amounts of unstructured input data and autonomously identify relevant information about tasks and workers.
In this regard, \LLMagent provides unprecedented opportunities to use comprehensive models of workers that integrate professional, psychological and social characterizations. Considering all the above aspects yields an in-depth view of employees and allows us to adopt a human-centric perspective in formulating personnel problems. For example, to evaluate the fit between a worker and a task, our agent checks if the worker possesses the required practical competences, whether the task aligns with his/her preferences and possible past interactions between the worker and the client requesting the task. If the task requires multiple workers, the agent also takes into account the compatibility and differences between team members, so to create a united team and promote diversity.
In contrast to other automated staffing software, \LLMagent uses natural language to ask humans for additional information and provide explanations. As a result, the process becomes more intuitive, understandable and manageable by humans, improving user experience. Remarkably, by analyzing human feedback, reflecting on the outcome of past actions and asking questions, our LLM agent \textit{actively learns and improves over time}.
Overall, \LLMagent combines algorithmic efficiency with LLM-level reasoning, solving the staffing and profiling problems in a fully automated, scalable and human-friendly way.

\begin{figure}[htbp]
  \centering
  \includegraphics[width=0.72\linewidth]{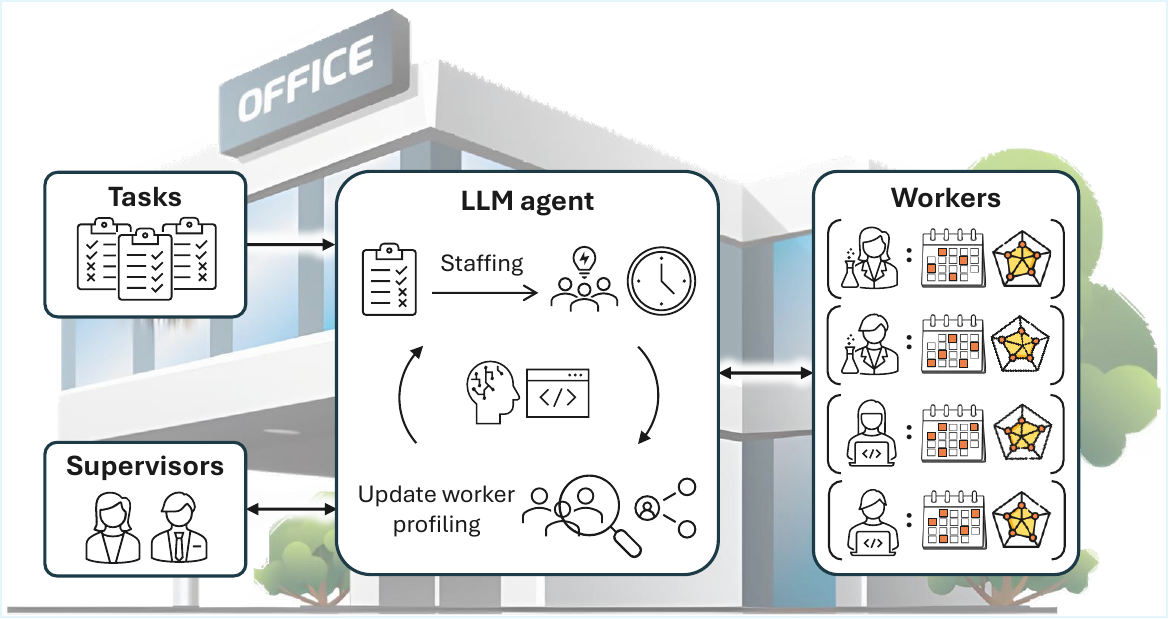}
  \caption{Our LLM agent combines automated staffing (task assignment and scheduling) with profiling (estimation of workers' attributes) to improve both over time. The agent receives orders from supervisors and directly interacts with workers, proposing schedules and analyzing their feedback.}
\end{figure}

\textbf{Contributions.}
Our contributions can be summarized as follows.
\begin{itemize}
    \item We formulate the novel problem of joint staffing and profiling with humans-in-the-loop, in which staffing and worker profiling are in symbiotic relationship. The humans-in-the-loop aspect is twofold: on the one hand, humans can decide whether to accept or reject schedules and, on the other hand, they generate a flow of sequential information relevant for the purposes of staffing and profiling.
    \item We propose \LLMagent, an LLM agent that solves the above problem, providing automated task staffing and lifelong worker profiling. Given a work environment with heterogeneous workers, our LLM agent optimizes the composition of the team assigned to each of the pending tasks according to user-defined optimization objectives, that can be a mix of analytical utility functions and textual criteria. The agent directly interacts with workers, proposing schedules and asking for necessary information. By analyzing human feedback and other data obtained as a by-product of staffing, the agent continuously estimates latent features of workers, such as their skills and personal preferences. This allows the agent to learn and improve its staffing performance over time. Using a combination of algorithms, LLM prompting and a dedicated scheduler, \LLMagent guarantees robustness, flexibility and interpretability.
\end{itemize}

%%%%%%%%%%%%%%%%%%%%%%%%%%%%%

\section{Problem Formulation}

In this section we present the main problem addressed by our proposed LLM agent, \LLMagent, namely joint staffing and worker profiling. To formulate the problem, we first introduce a general model for work environments that follow task-based work approaches, i.e., where employees are assigned to limited-time tasks or temporary project teams based on need.

\subsection{Work Environment}
We start by presenting a model that can describe structured workplaces ranging from public hospitals to consulting companies. The model has three types of elements, namely tasks, workers and supervisors. Tasks and workers are represented as generic data structures with customizable attributes, making the abstraction arbitrarily accurate and adaptable to changing requirements.

\textbf{Tasks}.
Table \ref{table:task_attributes} shows a possible characterization of tasks in which, among other things, we distinguish between full-time and part-time tasks. Full-time tasks monopolize workers for their entire duration and cannot be scheduled in overlapping time intervals. Part-time tasks let workers self-manage their time in a flexible way provided that they achieve the specified goals, and can be scheduled on top of other tasks.
We specify right away that the framework proposed in this article is highly generic and not limited to a specific type of tasks. Changes in the task template only affect the implementation of \LLMagent, leaving its high-level functioning unchanged. For example, the LLM agent we tested in our experiments can be easily adapted to support tasks that require specific resources subject to availability constraints, or multimodal task descriptions.
%In what follows we focus on tasks with attributes as in Table \ref{table:task_attributes}. However, the framework proposed in this paper is more generic and supports tasks with attributes of different types and formats. For example, one may want tasks to require specific resources subject to availability constraints, or task descriptions to be multimodal. These needs can be met with small code adjustments and choosing the right building blocks, while keeping the high-level functioning of the agent unchanged.

\begin{table}[h!]
\begin{tabularx}{\textwidth}{@{}l X@{}}
Attribute & Attribute description\\
\midrule
Arrival time & Timestamp when the task was added to the list of pending tasks.\\
Description & Unstructured natural language description of the task.\\
Priority level & Relative importance and urgency of the task.\\
Required roles & Professional roles required by the task and their number.\\
Duration & Estimated time needed to complete the task.\\
Workload & Effort required to complete the task by each worker.\\
Team constraints & Workers necessarily assigned to or exempted from the task.\\
Deadline & Timestamp by which the task must be completed.\\
Timing & Full-time commitment or self-managed part-time.\\
\bottomrule
\end{tabularx}
\vspace{3pt}
\caption{Task template used in this work.}
\label{table:task_attributes}
\end{table}

\textbf{Workers}.
Table \ref{table:worker_attributes} shows our structured representation of workers, that allows for informed task assignment and effective worker profiling. Also in this case, the proposed template can be modified to include additional details and adequately capture worker heterogeneity. 

\begin{table}[h!]
\begin{tabularx}{\textwidth}{@{}l X@{}}
Attribute & Attribute description\\
\midrule
Professional role & Job title and seniority.\\
Salary & Compensation of the worker per time unit.\\
Calendar & Ongoing and future tasks assigned to the worker.\\
Work history & Tasks completed by the worker and their outcomes.\\
Work capacity & Maximum daily workload the worker can manage.\\
Hard skills & Graded technical task-related competences.\\
Soft skills & Graded interpersonal and behavioral skills.\\
Task preferences & General preferences and aversions about tasks.\\
Teammate preferences & Compatibility with work colleagues.\\
\bottomrule
\end{tabularx}
\vspace{3pt}
\caption{Worker template used in this work.}
\label{table:worker_attributes}
\end{table}

We briefly illustrate the modeling of workers' skills and preferences, that in this paper are assumed to be unknown and need to be estimated.
We model skills as key-value items, where keys are skill names expressed in natural language and values are proficiency levels. Proficiency can be measured following any convention that an LLM can understand, such as letter grade, star rating, or qualitative scales like ``No competence, Beginner, Intermediate, Advanced, Expert” commonly used in skill matrices. Expressing the degree of proficiency using levels rather than arbitrary sentences has multiple advantages: it allows reducing the ambiguity of evaluations, easily tracking the evolution of competences over time, and converting levels into numerical values that can be used by algorithms.
We also model preferences as key-value items, where keys are names of work colleagues in the case of teammate preferences, and tags in the case of task preferences. Tags are short sentences that can address any aspect of a task, such as the client requesting the service or specific details mentioned in the task description. Values denote the direction and level of the preference, for example ``Strong aversion, Slight aversion, Neutral, Slight preference, Strong preference”.

\textbf{Supervisors}.
Placed at a higher level in the decision-making hierarchy, supervisors may be project managers, human resources managers or other figures responsible for coordinating and managing workers.
Supervisors direct staffing activities by setting the objectives to be optimized and their relative importance, adding task-specific constraints that the final schedule must satisfy, handling workers' requests and reviewing schedules. For example, supervisors interested in staff training may require that a worker is assigned certain tasks to develop new skills. Conversely, when workers ask to be exempted from tasks they have been assigned to, supervisors can accommodate workers' requests by adding exclusion constraints and ordering rescheduling. The task template in table \ref{table:task_attributes} already supports mandatory assignment and exclusion constraints, allowing supervisors to enforce or prohibit the assignment of tasks to the specified workers.

\subsection{Joint Staffing and Profiling Problem}
\label{sec:problem_formulation}

We now provide a mathematical formulation of joint staffing and profiling, starting with a formal statement of staffing problems. In most work environments staffing is a recurring problem that has to be solved periodically or in an event-based manner, for example when new tasks arrive. Considering a discrete time scale, we define staffing at time $k$ as a combinatorial and constrained optimization problem with the following inputs:
\begin{itemize}
    \item $T_k$, the list of pending tasks at time $k$.
    \item $W_k$, the set of workers at time $k$. Each element of this set describes the state of a worker, consisting of updated information about the worker that do not need to be estimated, such as professional role, work capacity, salary and task calendar.
    \item $\left\{ u_i(\cdot) \right\}_{i \in [1, n]}$, the set of optimization objectives to be maximized. Each $u_i$ is a scalar utility function modeling any kind of decision criterion. For example, $u_1$ may compute a mathematical quantity, $u_2$ may rate the satisfaction of a decision criteria expressed using natural language, and the evaluation of $u_3$ may involve the analysis of images or other files. Compared to similar problems proposed in the literature, 
    %
    %our heterogeneous formulation is the first of its kind.
    our formulation is the first that allows for this mix of heterogeneous optimization objectives.
    \item $\left\{ c_i \right\}_{i \in [1, n]}$, a set of coefficients expressing the relative importance of each optimization objective. Setting negative coefficients is equivalent to minimizing the corresponding objectives instead of maximizing them.
\end{itemize}

\begin{definition}[\itshape Schedule]
A solution to the staffing problem, called schedule for simplicity, is a map $s: t \in T_k \to \bigl( \text{team} \subseteq W_k, \text{time interval})$ assigning to each pending task a pair that specifies a subset of workers and a time scheduling.
Let $x_{wt} \in \{0, 1\}$ be a binary variable indicating whether worker $w$ is assigned to task $t$.
Let $\mathcal{I}_k = \bigl\{ [\alpha, \beta) \ | \ \alpha, \beta \in \mathbb{N},\ k \leq \alpha < \beta \bigr\}$ be the set of time intervals starting from time $k$, possibly up to a given time horizon.
Then, a schedule can also be represented as a tuple $s = \bigl( \{ x_{wt} \}_{w \in W_k, t \in T_k}, \{ [\alpha_t, \beta_t) \}_{t \in T_k} \bigr)$.
\label{def:feasible_schedule} 
\end{definition}
A schedule that satisfies all the existing constraints, which depend on the way tasks and workers are defined, is called a \textit{feasible schedule}. The set of feasible schedules at time $k$ is denoted by $\mathcal{S}_k \subseteq T_k \times 2^{|W_k|} \times \mathcal{I}_k$.
The goal of staffing is to determine the feasible schedule $s_k$ that finds the approval of all the workers involved and maximizes the optimization objectives proportionally to their relative importance. This proportional maximization is done by maximizing a value function $V(\cdot)$ that aggregates the individual objectives, such as a simple weighted average with weights $\left\{ c_i \right\}_{i \in [1, n]}$.
When tasks have attributes as in Table \ref{table:task_attributes}, we can formulate the staffing problem at time $k$ as follows. Below, $\mathbb{1}(\cdot)$ is the indicator function, \code{days} is the set of future workdays, and each workday $d \in \code{days}$ is a collection of time indices.
\begin{subequations} \label{eq:MIP}
\begin{alignat}{8}
    & s_k = \argmax_s \; V \left( u_1 \left(s\right), \dots, u_n \left(s \right), \left\{ c_i \right\}_{i \in [1, n]} \right) \label{eq:aggregation_function_maximization} \\
    & \text{such that } \forall t \in T_k,\  \forall w \in W_k,\ \forall d \in \code{days},\ \forall \text{ time index } h > k: \nonumber \\
    & \beta_t = \alpha_t + t.\code{duration} \label{eq:constraint_duration} \\
    & \beta_t \leq t.\code{deadline} \label{eq:constraint_deadline} \\
    & \sum_{w \in W_k} x_{wt} \cdot w.\code{role} = t.\code{requirements} \label{eq:constraint_requirements} \\
    & \sum_{h \in d} \sum_{t \in T_k} x_{wt} \cdot \mathbb{1}(\alpha_t \leq h < \beta_t) \cdot \frac{t.\code{workload}}{t.\code{duration}} \leq w.\code{work\_capacity} \label{eq:constraint_workload} \\
    & \sum_{t \in T_k} x_{wt} \cdot \mathbb{1}(\alpha_t \leq h < \beta_t) \cdot \mathbb{1}(t.\code{timing} = \code{full\_time}) \leq 1\label{eq:constraint_full_time} \\
    & x_{wt} \in \{0, 1\},\ \alpha_t \in \mathbb{N},\ \beta_t \in \mathbb{N},\ \alpha_t \geq k \nonumber
\end{alignat}
\end{subequations}

The constraints in the above a mixed integer program ensure the following feasibility conditions: \textit{(i)} the time interval in which each task is scheduled is equal to the duration of the task \eqref{eq:constraint_duration}, \textit{(ii)} all tasks are completed within their deadline \eqref{eq:constraint_deadline}, \textit{(iii)} the composition of the team assigned to each task satisfies the requirements of the task, e.g. professional roles are present in the required number \eqref{eq:constraint_requirements}, \textit{(iv)} the total daily workload assigned to each worker does not exceed the daily work capacity of the worker \eqref{eq:constraint_workload}, \textit{(v)} at each time instant, each worker is not assigned to multiple full-time tasks \eqref{eq:constraint_full_time}. Team constraints can be enforced by fixing the value of specific $x_{wt}$.

\algorithmstyle{ruled}
\begin{algorithm}
\DontPrintSemicolon
\SetNoFillComment
\SetArgSty{textnormal}
\SetCommentSty{mycommentfont}
$T_0,\ s_0,\ W_0,\ \hat{\theta}_0 \leftarrow$ initial set of tasks, schedule, workers' states and estimated attributes.\par
\vspace{4pt}
\For{each time step $k = 1, 2, \dots$}{
    \tcp{State = [tasks, workers]}
    $T_k, W_k, f_k^{PF}, f_k^{PR} \leftarrow$ \code{state\_transition}$(T_{k-1}, W_{k-1}, s_{k-1})$\par
    \tcp{Solve the current staffing problem}
    \While{\code{True}}{
        $s_k \leftarrow$ \code{schedule}$( T_k, W_k$ \hlfancy{white}{$|\ \hat{\theta}_{k-1}$}$)$\par
        \code{human\_decision}, $f_k^{TP} \leftarrow$ \code{propose$(s_k)$}\par
        \If{\code{human\_decision == accept}}{
            $T_k, W_k \leftarrow$ \code{confirm$(s_k)$}\par
            \textbf{break}
        }
    }
    \tcp{Use feedback to update the estimated workers' attributes}
    \hlfancy{white}{ $\hat{\theta}_k \leftarrow$ \texttt{update\_profiling}$\left(  \mathcal{F}_{1:k} \right)$ }
}
\caption{Joint staffing and profiling (sequential)}
\label{alg:staffing_profiling_formulation}
\end{algorithm}

Algorithm \ref{alg:staffing_profiling_formulation} describes a basic pipeline that integrates staffing and profiling.
The state transition function \code{state\_transition$(\cdot)$} models the progression and end of tasks, the arrival of new tasks, changes in the set of available workers and other dynamics of interest.
The presence of humans-in-the-loop requires that staffing problems be solved iteratively: for example, workers can reject tasks assigned to them provided they obtain permission from their supervisor. This results in the \code{while} loop in the algorithm.
The function \code{schedule$(\cdot)$} generates a feasible schedule $s_k$ given the current state $[T_k, W_k]$ by solving the maximization problem \eqref{eq:MIP}. The conditioning on the variable $\hat{\theta}_k$, which contains the current estimates of workers' attributes, indicates that profiling has a direct effect on staffing.
The function \code{propose$(\cdot)$} forwards the generated schedule to workers first, who provide feedback and can ask for changes in the schedule, and to supervisors then, who review the schedule and workers' requests.
Supervisors can either confirm the schedule by setting the variable \code{human\_decision} equal to \code{accept}, or update the optimization objectives and task-specific constraints to generate a new schedule.
The above procedure produces multiple types of human feedback, such as peer feedback $f_k^{PF}$, performance reviews $f_k^{PR}$ and feedback about task proposals $f_k^{TP}$. The set of all feedback collected at time $k$ is denoted with $\mathcal{F}_k$.

The optimization objectives can address specific attributes of workers, such as their proficiency levels in technical and interpersonal skills or their personal preferences, whose true value is not known in general and needs to be estimated. This leads us to formulate the worker profiling problem. Adopting the worker representation in Table \ref{table:worker_attributes}, in this work we aim to learn the proficiency levels of workers in both hard and soft skills and workers' preferences about tasks and teammates. Let the latent vector $\theta_k = \left[ \theta_k^1, \dots, \theta_k^{|W|} \right]$ represent the true values of workers' attributes at time $k$, where each sub-vector $\theta_k^w \in \Theta$ refers to a different worker $w \in W$ and $\Theta$ is the set of vectorized attribute values that can be chosen.
The profiling problem at time $k$ consists in obtaining an updated estimate $\hat{\theta}_k$ of the true attributes of workers through the analysis of the cumulative feedback $\mathcal{F}_{1:k}$ collected over time. From a high-level perspective, finding the estimator that is most coherent with the observed feedback can be framed as a maximum likelihood estimation problem. Assuming an unknown probability distribution $p(\cdot)$, the attributes that are most likely to have generated the observed feedback are obtained by maximizing the likelihood $L(\cdot)$:
\begin{equation} \label{eq:profiling_as_MLE}
    \hat{\theta}_k = \argmax_{\theta' \in \Theta^{|W|}} \left\{ L(\theta') = p(\mathcal{F}_{1:k} | \theta') \right\} .
\end{equation}

\begin{figure}[tb]
  \centering
  \includegraphics[width=0.7\linewidth]{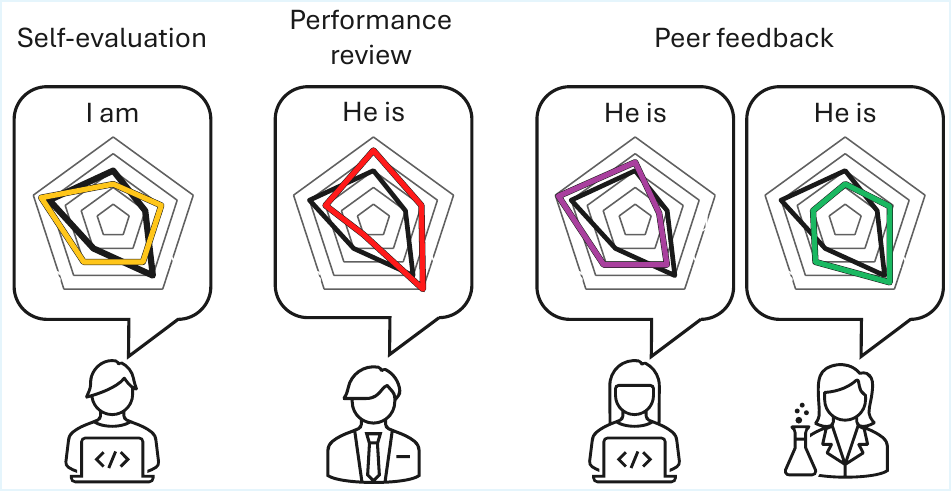}
  \caption{Visual representation of how people may rate the skills of a target worker. In each radar chart, the actual skill levels of the target worker is black, and the levels the speaker says the worker has is of a different color. Due to cognitive biases, personal interests and other phenomena, the statements from co-workers, supervisors, and even the target worker himself are usually inaccurate.
  }
  \label{fig:noise_and_bias}
\end{figure}

From \eqref{eq:profiling_as_MLE} it is clear that the quality of the available raw information is the limiting factor of the achievable profiling performance. 
Given the importance of $\mathcal{F}_{1:k}$ in the above formulation, we take a closer look at the sources of raw information on which profiling is based. Examples of these sources are reports from supervisors and feedback from work colleagues, namely external and possibly contrasting opinions that are naturally distorted by cognitive biases and social dynamics. Even direct self-evaluations from workers cannot be fully trusted, as their accuracy depends on the self-awareness, objectivity and honesty of workers.
This is visually represented in Figure \ref{fig:noise_and_bias}.
Taking into account these aspects, we propose the feedback generation model \eqref{eq:feedback_model}, that uses the concepts of observer and observation. We call observers people who can provide information about workers, such as supervisors, clients, work colleagues and workers themselves, and use $O_k$ to denote the set of observers at time $k$.
We define a skill observation as a tuple $(w,\ o,\ \mathscr{s}, \ \theta_k^{w}[\mathscr{s}])$, where $w$ is the target worker, $o$ is the observer providing the observation, $\mathscr{s}$ is the name of the skill, and $\theta_k^{w}[\mathscr{s}]$ is the corresponding competence level of the worker at time $k$. Similarly, a preference observation is a tuple $(w,\ \mathscr{p}, \ \theta_k^{w}[\mathscr{p}])$, where $w$ is the target worker, $\mathscr{p}$ is the name of the preference, and $\theta_k^{w}[\mathscr{p}]$ is the corresponding preference level. In preference observations the observer field is omitted and is implicitly $o=w$, as we assume that the only observer of a worker's preferences to be considered for profiling purposes is the worker himself.

\begin{subequations} \label{eq:feedback_model}
  \begin{equation} \label{eq:feedback_model_a}
  \begin{split}
     \mathcal{F}_k \ni f \sim (\code{to\_text} \circ \code{mask}) \bigl( \mathcal{O}_k^{\text{skills}}
     \ \cup \ \mathcal{O}_k^{\text{preferences}} \bigr)
  \end{split}
  \end{equation}
  \begin{equation} \label{eq:feedback_model_b}
  \begin{split}
     \mathcal{O}_k^{\text{skills}} = & \bigl\{ \bigl( w, \ o, \ \mathscr{s}, \ \theta_k^{w}[\mathscr{s}] + b^{o,w,\mathscr{s}} + v \bigr) \ | \  \mathscr{s} \in \text{skills}(w), \ w \in W_k , \ o \in O_k, \ v \sim \mathcal{N}(0, \sigma_v^2) \bigr\} \\
     & \text{where each} \ b^{o,w,\mathscr{s}} \sim \mathcal{N}(0, \sigma_b^2)
  \end{split}
  \end{equation}
  \begin{equation} \label{eq:feedback_model_c}
  \begin{split}
  \mathcal{O}_k^{\text{preferences}} = \bigl\{ \bigl( w, \ \mathscr{p},\ \theta_k^{w}[\mathscr{p}] \bigr) \ | \ \mathscr{p} \in \text{preferences}(w), \ w \in W_k \bigr\}
  \end{split}
  \end{equation}
\end{subequations}

In the model \eqref{eq:feedback_model}, \code{mask}$(\cdot)$ takes a set of attribute observations as input and returns a small subset of them based on the type of feedback to be generated. The transformation \code{to\_text}$(\cdot)$ maps a set of attribute observations to a natural language instance, and is typically stochastic, lossy and not invertible.
The composition of these two operators in \eqref{eq:feedback_model_a} indicates that feedback is a natural language encoding of partial observations of the skills and preferences of workers. The skill observation set $\mathcal{O}_k^{\text{skills}}$ defined in \eqref{eq:feedback_model_b} contains all the possible observations of workers' skills at time $k$, considering all the combinations of skills, target workers and observers. To model the subjectivity and stochasticity of human judgment, the competence levels in these observations are corrupted by random noise $v$ and bias $b^{o,w,\mathscr{s}}$, where the bias is specific for each observer-worker-skill triplet. The preference observation set $\mathcal{O}_k^{\text{preferences}}$ defined in \eqref{eq:feedback_model_c} contains all the possible observations of workers' preferences. Since these observations come from workers themselves, we assume workers do not lie about their preferences, and thus the observations in this set contain true preference levels.

For explanatory clarity, in Algorithm \ref{alg:staffing_profiling_formulation} the two maximization problems \eqref{eq:MIP} and \eqref{eq:profiling_as_MLE} are optimized sequentially. Instead, \LLMagent implements event-based profiling, analyzing new feedback as soon as it is collected to promote better data utilization.

%@NICOLO': MAGARI AGGIUNGI UNA FRASETTA SUL PARAGONE CON sysID + OPTIMAL CONTROL

%%%%%%%%%%%%%%%%%%%%%%%%%%%%%

\section{LLM Agent for Staffing and Profiling}
\label{sec:LLM_agent}

\begin{figure}[htbp]
  \centering
  \includegraphics[width=0.85\linewidth]{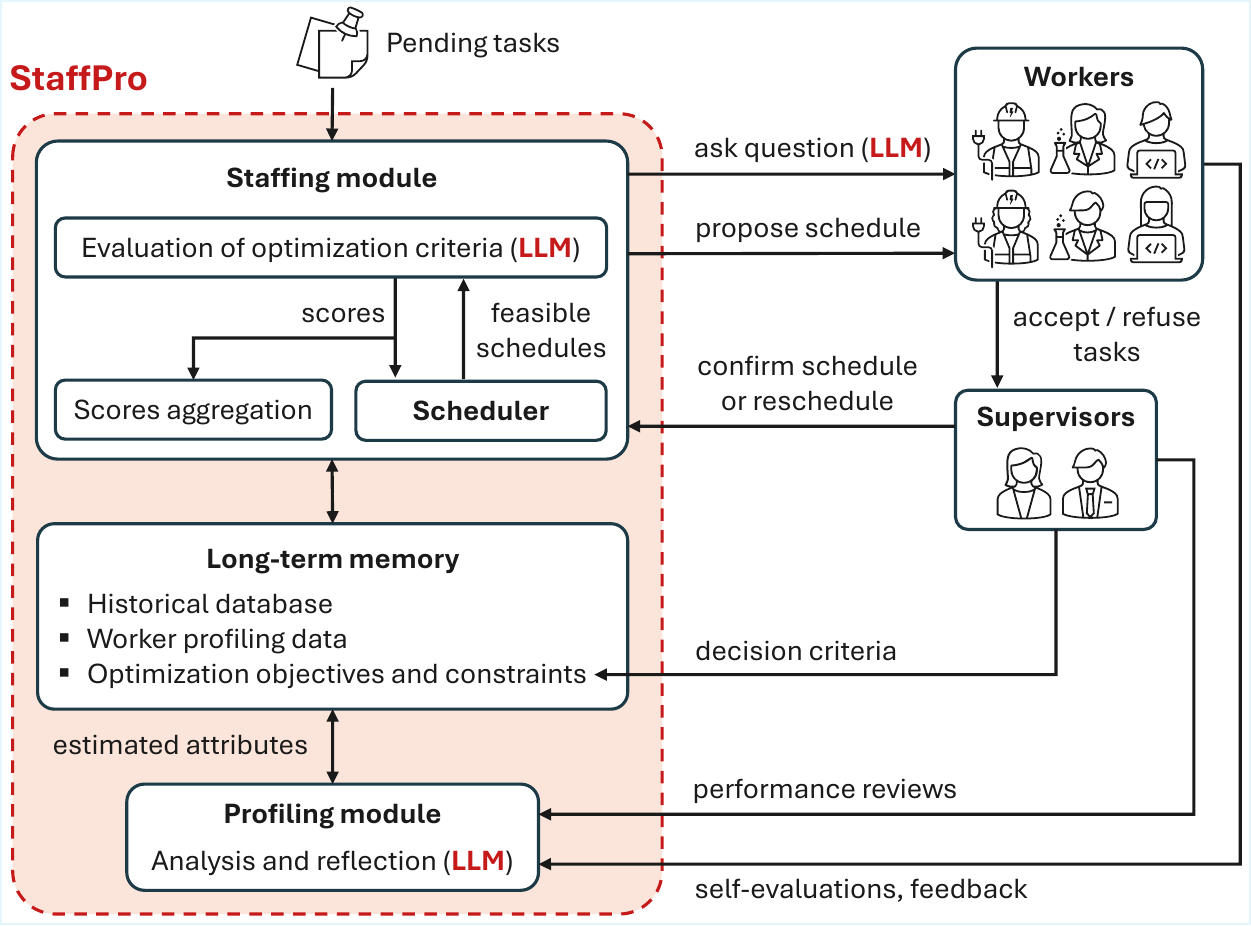}
  \caption{High-level representation of the proposed LLM agent: architecture, essential workflow and integration into the work environment.}
  \label{fig:LLM_agent_architecture}
\end{figure}

In this section we present \LLMagent, our automated solution to the joint staffing and profiling problem. \LLMagent is an LLM agent featuring two main modules: the staffing module generates schedules for pending tasks and proposes them to workers, the profiling module analyses the data generated during work processes to estimate the characteristics of workers. Both modules can query and update the long-term memory of the agent, which stores profiling data, data about past tasks, optimization objectives and other information relevant for staffing. Modules and memory are tied together by an API which triggers procedures based on events, connects the LLM agent to the external environment and allows interaction between agent and humans.
Importantly, we simply rely on the built-in knowledge of an arbitrary off-the-shelf LLM, without the need to fine-tune a model, and treat it as a building block that can be replaced when better models become available.
Figure \ref{fig:LLM_agent_architecture} provides a high-level representation of the internal architecture of the LLM agent and shows how the latter is integrated into the overall workflow.
Below we describe the overall functioning of \LLMagent, focusing on staffing in Section \ref{sec:LLM_agent_staffing} and on profiling in Section \ref{sec:LLM_agent_profiling}. The generation of schedules involves using a dedicated algorithmic scheduler, which is treated as a black box. To make our work self-contained, in Appendix \ref{appendix:scheduler} we present a possible scheduling algorithm based on a multi-path conditional planning strategy. 

\subsection{Staffing Module} \label{sec:LLM_agent_staffing}
The staffing module groups together the components of the agent dedicated to solving staffing problems. This module generates candidate schedules for the current set of pending tasks and proposes them to workers until a schedule is accepted. Below we illustrate the overall functioning of the module and how staffing problems are addressed.

\textbf{Batch staffing.} The staffing module is activated when 
%
%a new task arrives or, preferably, 
%
the queue of pending tasks $T_k$ reaches a given number of elements. In fact, tasks should not be processed individually as soon as they arrive, as this would result in greedy task assignments that do not consider subsequent tasks. Rather, tasks should be jointly scheduled in batches so to obtain better schedules that optimize multiple tasks.

\textbf{Not just an LLM.} The staffing problem formulated in Section \ref{sec:problem_formulation} is combinatorial, constrained, and aims to maximize an arbitrary number of optimization objectives, partly analytical and partly textual, proportionally to their relative importance. Simply prompting the LLM to generate a schedule given tasks, workers, constraints, optimization objectives and importance coefficients would hardly produce good results. In fact, while LLMs exhibit impressive reasoning and code generation capabilities, they are also prone to errors.
%
%and not necessarily the best choice. 
Instead, \texttt{StaffPro} makes the LLM interact with other tools, including a scheduler that handles constraints algorithmically to ensure they are met, and a set of functions to evaluate standard analytical optimization objectives. The main idea is to combine the strengths of different approaches and resort to the LLM only when there are no better alternatives. Consistent with this line of thinking, we do not take advantage of the real-time code generation capabilities of LLMs, preferring a more conservative, structured, and reliable approach.

\textbf{Decision criteria.} The optimization objectives and their importance levels are set by supervisors, who can choose from a list of options and add new custom objectives. Optimization objectives are typically task-agnostic and long-term, and are thus stored in the long-term memory, but they can be changed at any moment. Additionally, the agent may be given task-specific constraints that the schedule to be generated must satisfy. For example, supervisors may enforce or prohibit the assignment of a task to some workers, or require two tasks to be assigned to non-overlapping sets of workers. In any case, constraints must be supported by the scheduler, as specified below.

\textbf{Finding the right schedule.}
%Identifying the best schedules can be broken down into two main steps: generating feasible schedules and evaluating optimization objectives. The first duty is taken care of by a dedicated scheduler, such as a standard MIP solver applied to \eqref{eq:MIP}, ensuring constraint satisfaction by design. We consider the scheduler as a black box and allow for arbitrary scheduling algorithms satisfying the feasibility conditions, proposing a possible scheduler in Appendix \ref{appendix:scheduler}. Regarding the evaluation of optimization objectives, our approach is based on obtaining numerical scores that represent the degree to which each objective is satisfied by a given candidate. Consistent with the problem formulation in Section \ref{sec:problem_formulation}, in our system each objective is represented by a scalar function $u_i(\cdot)$ that allows to evaluate it. These functions can be implemented using standard algorithms or also leveraging the LLM to partly or entirely evaluate the objective. Our prompting approach is memory-less, includes minimal in-context examples and imposes structured output formats to ease memory management. As an example, below we show a prompt template to rate the suitability of different workers for a task based on their technical skills. The prompt makes the LLM generate intermediate quantities that are then automatically validated and used to compute the value of an objective. 
Identifying the best schedules requires solving the scheduling problem \eqref{eq:MIP} or a similar one, depending on the specific application. This type of combinatorial and constrained problem is difficult to solve exactly, and is often addressed by resorting to heuristic methods that provide cheaper approximate solutions. To keep our design flexible, we do not impose a specific solution method: we consider the scheduler as a black box, allowing the scheduling problem to be solved in any way, as long as the feasibility constraints are satisfied.
In our simulations we use a new scheduling algorithm proposed in Appendix \ref{appendix:scheduler}, which represents a minor contribution of this work. Regardless of the algorithm chosen, solving the scheduling problem involves evaluating the optimization objectives. As described in Section \ref{sec:problem_formulation}, in our system each objective is represented by a scalar function $u_i(\cdot)$ that allows to evaluate it, returning a numeric score that represents the degree to which the objective is satisfied by a given candidate. These functions can be implemented using standard algorithms or also leveraging the LLM to partly or entirely evaluate the objective. Our prompting approach is memory-less, includes minimal in-context examples and imposes structured output formats to ease memory management. As an example, below we show a prompt template to rate the suitability of different workers for a task based on their technical skills. The prompt makes the LLM generate intermediate quantities that are then automatically validated and used to compute the value of an objective. 

\begin{mybox}{\small Example of prompt for LLM-powered criteria evaluation}
\small
Consider this task: <task description>\newline
We want to determine how well each of the following workers is suited to the task.\newline
<workers' estimated profiling data>\newline
These are the possible skill levels and their scores: <qualitative skill scale>\newline
Assign a score to each worker and add a short justification. First analyze the task description, identifying which hard skills are required for each job role to perform its part of the task. Then, assign a score to each worker based on their level of proficiency in the corresponding hard skills.
Answer with a list of <number of workers> Python objects, one for each worker, with attributes `worker\_id', `score' and `justification'.\newline\newline
Example:\newline
Task: `... the Operations Consultant will optimize the supply chain to minimize waste and improve on-time delivery to construction sites ...'\newline
Worker: Sarah [Professional role: Operations Consultant. Hard skills: Lean Manufacturing (level = Beginner); Supply Chain Management (level = Advanced); Process Optimization (level = Intermediate)]\newline
Your output: [worker\_id = `Sarah', score = 6, justification = `The hard skill required by the Operations Consultant for the task is Supply Chain Management, in which Sarah has Advanced-level competence.']
\end{mybox}

To ensure the correct functioning of the agent and improve its performance, each function $u_i$ should be coded in advance so to limit the use of LLM to when it is strictly necessary, and in such cases carefully craft and test prompts. For objectives that are not associated with a dedicated evaluation function, the agent compiles a generic prompt containing the task assignments to be evaluated, relevant profiling data, and the optimization objective as provided by the human user, leaving the evaluation of the objective entirely up to the LLM.
Once all optimization objectives have been evaluated, the resulting numerical scores and their importance levels $c_i$ are fed to the aggregation function $V(\cdot)$, which provides an overall evaluation of candidate schedules and allows maximizing optimization objectives proportionally to their importance. Following \eqref{eq:aggregation_function_maximization}, the schedule chosen to be proposed to workers is the one maximizing $V(\cdot)$. Alternatively, if the scheduler used is capable of generating multiple schedules, one can replace $\argmax$ in \eqref{eq:aggregation_function_maximization} with a argtop-$K$ operator to select the $K$ schedules with the highest aggregate score. Importantly, the whole generation process can be traced and sees a clear separation of roles between scheduler, LLM and scripts, ensuring interpretability.

\textbf{Criteria evaluation via LLMs.} In our approach, the LLM serves as generic evaluation function mapping pairs (candidate, objective) to scalar values inside a specified range. For example, we use LLMs to measure the satisfaction of qualitative criteria using a qualitative scale, which is then converted to a numerical scale to obtain a score. This allows exploiting the semantic understanding and reasoning capabilities of LLMs, that can identify similarities between concepts expressed using different words, make deductions and generalize. Moreover, the LLM is instructed to always justify its choices, promoting explainability.

\textbf{Informed decision-making.} The evaluation of certain optimization objectives may require knowing specific attributes of workers. If such information is stored in the long-term memory, the staffing module will simply access it. In particular, if the query concerns the work history of the worker, only a subset of relevant tasks completed by the worker are retrieved from the memory. The selection is based on the cosine similarity of the embeddings of the task descriptions, computed between past tasks and a target task to be assigned. Especially at the beginning, the information available to the agent may not be sufficient to evaluate certain optimization objectives. In some cases the lack of information is detected algorithmically, in others the LLM is prompted to return certain values to signal that the information provided is incomplete. Depending on the optimization objective, the agent either uses default values for unknown attributes or directly asks the worker for the required attribute. For example, the agent may ask a worker about his/her competence in a specific programming language, and then save this information in the database. When interacting with humans, \LLMagent may also calibrate his writing and speaking style based on the psychological profile of the target worker, improving the effectiveness of communication and the way interactions are felt.

\textbf{Getting approval.} Once one or multiple candidate schedules have been generated, the staffing module proposes them to workers, either by broadcasting the whole schedule to all the workers involved, or by separately asking workers about specific tasks assigned to them. Schedules are proposed once at a time in decreasing order of aggregated score, making the process iterative. Workers are asked to accept or reject the tasks assigned to them, and are invited to provide optional feedback and suggestions to improve subsequent assignments. If a schedule finds the approval of all the workers involved the process ends successfully: all the tasks included in the schedule are removed from the queue of pending tasks and are added to the calendars of the workers. Otherwise, task rejections are reviewed by supervisors and the next candidate schedule is proposed to workers. If none of the schedules satisfies the workers unanimously, supervisors are asked to confirm one of the rejected schedules anyway or to invoke rescheduling, possibly updating the optimization objectives and adding constraints to ensure that specific workers’ needs are satisfied. In any case, the rejection motivations and suggestions provided by workers are stored and analyzed by the profiling module, so to update the estimates of workers' preferences and improve future schedules.

\textbf{Helping supervisors to decide.} In case multiple rescheduling rounds are performed due to task refusals, \LLMagent can provide supervisors with simple statistics showing how refusals affected the quality of the final schedule. One way to summarize the schedule evolution is to compare the initial schedule with the final one. This can be done by listing pending tasks and highlighting which tasks are included in each plan, the corresponding scheduling times and differences in terms of scores associated with the optimization objectives. Supervisors can use this simple report to judge whether the task refusals that led to the final plan are acceptable, especially when the final plan is considerably worse than the initial one. For example, the final plan may not include some high-priority tasks that were present in the initial plan, or may result in some workers working significantly more than others, thus being unfair.

\subsection{Profiling Module} \label{sec:LLM_agent_profiling}

Worker profiling aims to bridge the gap between abstract roles and real workers by estimating implicit attributes of workers, such as their proficiency levels in skills and personal preferences. This is taken care of by the profiling module, that analyzes the byproducts of staffing problems such as self-evaluations and feedback about task assignments, collecting evidence about the characteristics of workers and updating the memory of the agent. This process allows \LLMagent to accumulate knowledge over time, refine profiling estimates and improve the staffing performance.

\begin{figure}[htbp]
  \centering
  \includegraphics[width=\linewidth]{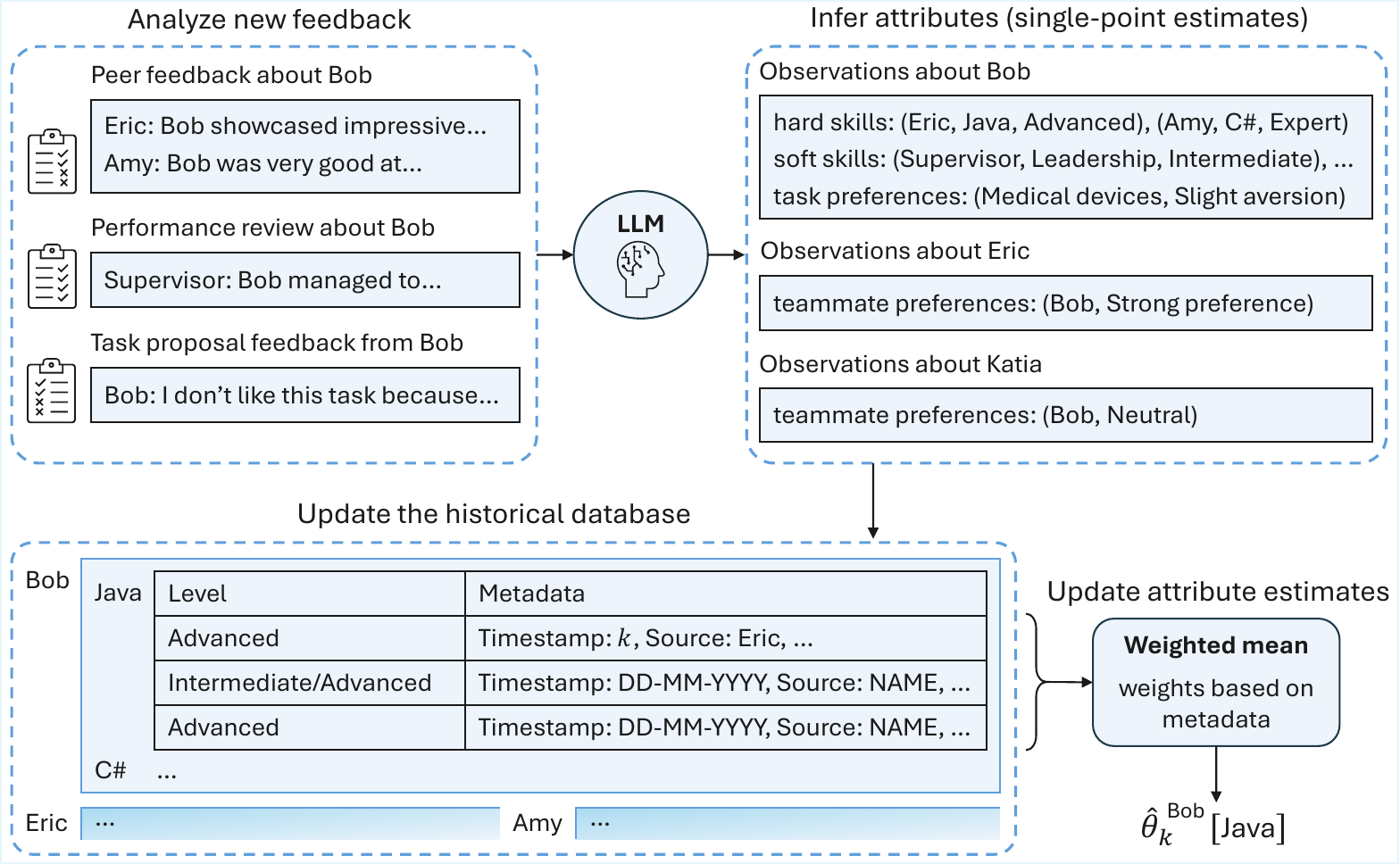}
  \caption{Our profiling pipeline consists in extracting observations about workers' attributes from the available information sources, and then combining these single-point estimates to obtain overall estimates. }
  \label{fig:profiling}
\end{figure}

\textbf{LLM-powered estimation.}
The main task of the profiling module is to fill in specific prompts with the yet-to-be-analyzed feedback in $\mathcal{F}_{1:k}$, and to update the long-term memory based on the output of the LLM. The profiling module activates as soon as a new piece of information is available, so that new knowledge can be exploited immediately. Depending on the type and source of the information, the module chooses between a set of different pipelines, each associated with specific estimation objectives, prompt templates and memory updates.
When analyzing third-party feedback or empirical evidence, such as peer feedback or performance reviews about the outcome of tasks, \LLMagent focuses on deducing skill levels. When the information comes directly from the worker to be profiled, as in the case of self-evaluations and feedback about task proposals, \LLMagent tries to infer both personal preferences and skill levels.
All prompts contain the new information to be analyzed and specific instructions to guide LLM inference. Additionally, some prompts can contain the current profiling data of the target worker and a toy input-output example to promote better output quality. LLM output compliance is ensured by constraining the output format, forcing the LLM to generate structured objects and specific data types that can be automatically validated, easily parsed and translated into memory updates.

\textbf{Recovering observations.}
In general, given an input feedback $f \in \mathcal{F}_{1:k}$ to be analyzed, \LLMagent instructs the LLM to extract skill observations $(w,\ o,\ \mathscr{s}, \ \ell)$ and preference observations $(w,\ \mathscr{p}, \ \ell)$, where $w$ is the target worker, $o$ is the observer providing the observation, $\mathscr{s/p}$ is the name of the skill/preference, and $\ell$ is the estimated level. This can be seen as trying to invert the operator \code{to\_text}$(\cdot)$ in \eqref{eq:feedback_model_a} to recover the observations that originated the analyzed feedback. Since the transformation \code{to\_text}$(\cdot)$ is typically non-invertible and the LLM can make interpretation errors, the retrieved observations can be different from the original ones are likely affected by some reconstruction noise.
When the LLM outputs an observation, the profiling module annotates it with the current timestamp, the type of information that was analyzed, a pointer to the original information, and a report including the prompt and the resulting output of the LLM. Timestamps are used to track the evolution of levels over time and give different importance to profiling data according to their recency. The reference to the original pieces of information ensures that everything can be traced, allowing humans to inspect and correct the deductions made by the LLM.
The annotated observation is then added to a dedicated data structure in the profile of the worker, creating a historical database where each attribute $\mathscr{a}$ to be estimated is associated with a list $R^{w,\mathscr{a}}$ of retrieved observations collected over time.

\textbf{Averaging over observations.}
As previously mentioned, retrieved observations are generally affected by noise, which can come from both the raw data and the deductions of the LLM. To smooth out this noise, we compute a weighted average of the estimated competence or preference levels contained in the individual observations, separately for each attribute. This is shown in \eqref{eq:weighted_average}, omitting the normalization factors for simplicity.
In the case of preferences, the weights are proportional to the recency of observations: in \eqref{eq:weighted_average_p}, $h$ is the timestamp associated to each observation and $\gamma \in (0,1]$ is the discount factor. This allows tracking time-varying preferences and addresses the fact that profiling is an online estimation problem.
In the case of skills, we also consider the source of the information: in \eqref{eq:weighted_average_s} each observation is weighted by an additional coefficient $\alpha^{o, w, \mathscr{s}}$ that determines how much the observer contributes to the average. The value of these coefficients should be chosen by supervisors, assigning greater importance to the more authoritative and impartial observers, so to mitigate the effect of observers' biases in \eqref{eq:feedback_model_b}.
The agent recomputes the weighted averages every time new observations are collected, resulting in a single and always updated estimate for each attribute, which is precisely the goal of worker profiling.
The above process implements the maximum likelihood estimation \eqref{eq:profiling_as_MLE} in two steps: we first analyze human feedback to extract observations, and then compute weighted averages to get the most plausible values of workers' attributes.

\begin{subequations} \label{eq:weighted_average}
  \begin{equation} \label{eq:weighted_average_p}
    \hat{\theta}_k^w[\mathscr{p}] \propto \sum_{(w, \mathscr{p}, \ell, h) \in R^{w,\mathscr{p}}} \gamma^{k-h} \ell
    \qquad \forall \mathscr{p} \in \text{preferences}(w)
  \end{equation}
  \begin{equation} \label{eq:weighted_average_s}
    \hat{\theta}_k^w[\mathscr{s}] \propto \sum_{(w, o, \mathscr{s}, \ell, h) \in R^{w,\mathscr{s}}} \gamma^{k-h} \alpha^{o, w, \mathscr{s}} \ell
    \qquad \forall \mathscr{s} \in \text{skills}(w)
  \end{equation}
\end{subequations}

%%%%%%%%%%%%%%%%%%%%%%%%%%%%%

\subfile{sections/experiments}

\subfile{sections/related_work}

\subfile{sections/conclusions}

\bibliographystyle{ieeetr}
\bibliography{references}

%%%%%%%%%%%%%%%%%%%%%%%%%%%%%

%\newpage
\appendix
\subfile{sections/scheduler}
\subfile{sections/experiments_appendix}

\end{document}

%% file: sections/experiments.tex
\section{Experimental Results} \label{sec:experimental_results}

Below we describe our experimental setup and present numerical results that demonstrate the advanced staffing and profiling capabilities of the proposed agent.
%The code of our experiments is publicly available at the GitHub repository \href{https://github.com/alemaritan/StaffPro}{StaffPro}. 

\subsection{Simulation Setup}

\textbf{Generating workers.}
We simulate a consulting work environment where workers are consultants with attributes as in Table \ref{table:worker_attributes}. We use a configuration script to set the number of workers, their seniority and professional role (e.g. IT Consultant, Legal Consultant, Business Consultant, etc.), the set of hard skills associated with each professional role, and the sets of soft skills and task preferences. Task preferences concern topics associated with tasks, for example cybersecurity, renewable energies or education. The remaining attributes are generated in a pseudo-random way: salary, work capacity and true proficiency levels in skills are random variables whose value depend on the seniority of the worker; true preference levels about tasks and teammates are randomly sampled from a discrete probability distribution. Preferences about teammates are asymmetrical, e.g. a worker may hate collaborating with a colleague but not vice versa. Consistent with \eqref{eq:feedback_model_b}, workers have biases that influence how they perceive and evaluate the skills of their colleagues.

\textbf{Generating tasks.}
We create tasks following the template in Table \ref{table:task_attributes}. We randomly choose the number of workers required for the task, their professional roles, a subset of their hard skills and some random topics. The task description is written by an LLM, which is prompted to generate a task requiring the specified workers and specializations and covering the given topics. To make the environment realistic, tasks are added to the queue of pending tasks at each time step according to a Poisson distribution, which is typically used to model packet arrival processes in telecommunications.

\textbf{Simulating feedback.}
To simulate workers seen from the perspective of \LLMagent, we need to generate the feedback that real workers would provide and the reports about their work performance. In most cases, we leverage an LLM to obtain realistic natural language feedback: the LLM is given a profiling statement containing the attributes of the workers to be impersonated, and specific instructions to guide the generation of the desired feedback. The provided workers' attributes are corrupted by bias and noise, consistent with the model \eqref{eq:feedback_model}.
As described in Appendix \ref{appendix:simulation_details}, we simulate four types of human feedback: self-evaluations, performance reviews, feedback about task proposals and peer feedback.

\textbf{Simulating supervisors.}
In our framework, supervisors are responsible for setting task-specific constraints, handling workers' requests and reviewing schedules. In our simulations we set constraints algorithmically, enforcing or prohibiting the assignment of each task to randomly chosen workers. For simplicity, we do not simulate workers' requests and always accept the first feasible schedule.

\textbf{Simulation size.} To talk numbers, we simulate a work environment with 2 junior and 2 senior workers for each of 5 professional roles, for a total of 20 workers. Each worker is associated with about 10 role-specific hard skills, 12 soft skills, 20 preferences about task-related topics and 19 preferences about work colleagues. We consider 7 qualitative skill levels, ranging from \doubleapex{No competence} to \doubleapex{Expert}, and 5 preference levels, going from \doubleapex{Strong aversion} to \doubleapex{Strong preference}. The standard deviation of the bias and noise used to generate human feedback according to \ref{eq:feedback_model} is $\sigma_b = \sigma_v = 1.5$. We set 9 staffing objectives to be optimized, listed in Appendix \ref{appendix:simulation_details}. To ease the generation of task descriptions by the LLM, we consider tasks for small teams with up to 3 workers. We generate 3000 distinct activities with durations ranging from 2 to 10 time steps. We use the scheduling algorithm presented in Appendix \ref{appendix:scheduler}, exploring $K=3$ scheduling paths in parallel.

\textbf{LLM model}. Our framework allows for any sufficiently advanced LLM that supports structured output generation. Since the LLM is used to evaluate some optimization objectives and analyze human feedback, the choice of the specific model directly affects the quality of both staffing and worker profiling. In our simulations we use the LLM \code{gemini-2.0-flash-001} and the embedding model \code{text-embedding-004}, both from Google.

\subsection{Performance Evaluation}

\begin{figure}
%\captionsetup[subfigure]{font=small} if you like to change caption style
     \begin{subfigure}[b]{0.49\textwidth}
         \includegraphics[width=\textwidth]{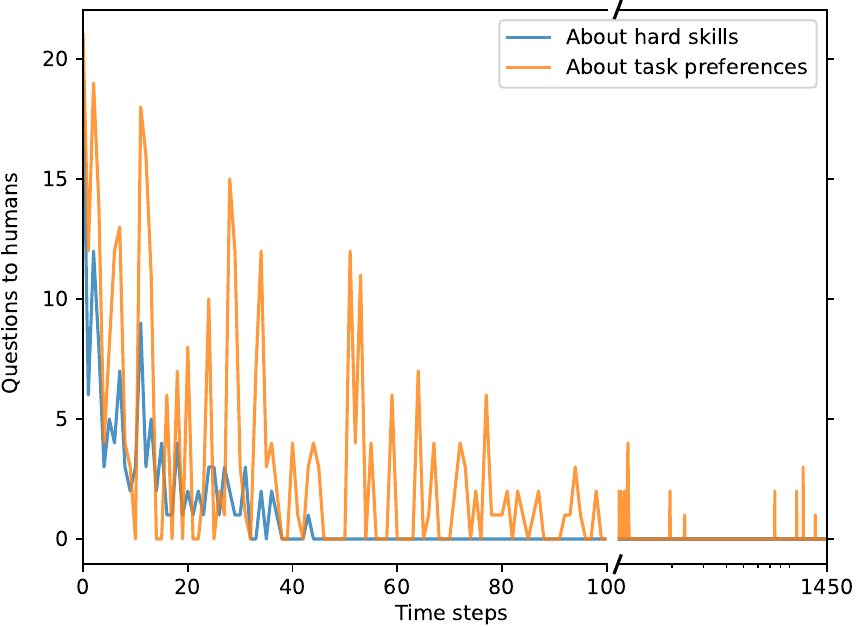}
         \caption[]{}
         \label{fig:human_requests}
     \end{subfigure}
     \hfill
     \begin{subfigure}[b]{0.49\textwidth}
         \includegraphics[width=\textwidth]{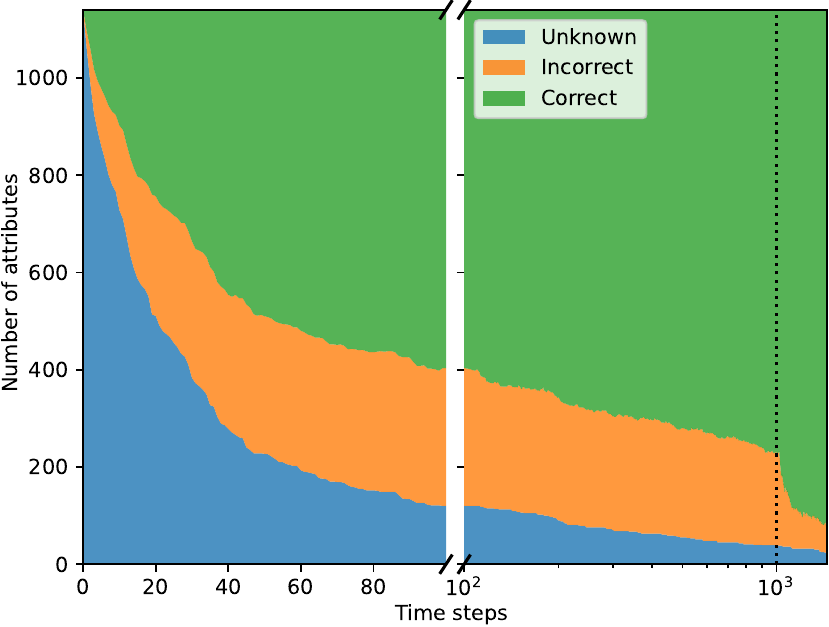}
         \caption[]{}
         \label{fig:attribute_knowledge_evolution}
     \end{subfigure}
     \vskip\baselineskip
     \begin{subfigure}[b]{0.49\textwidth}
         \includegraphics[width=\textwidth]{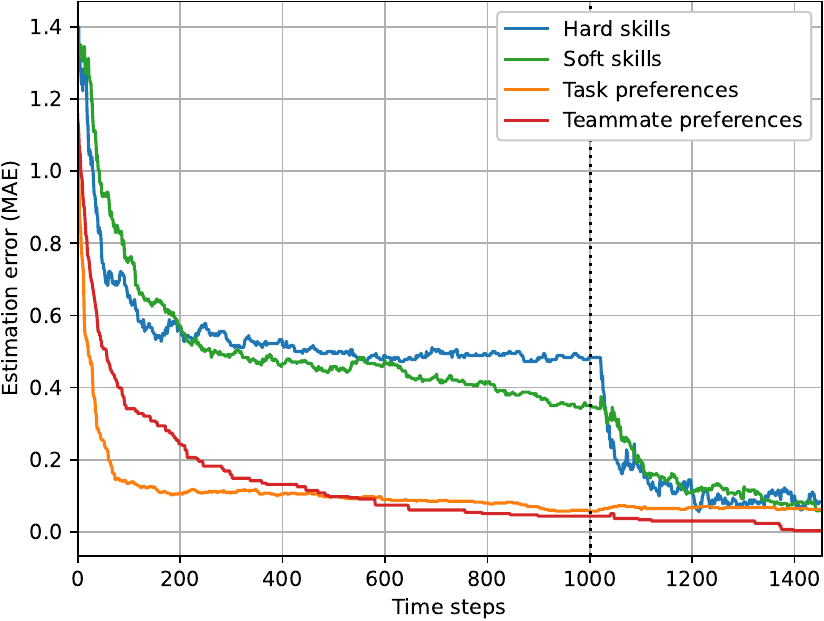}
         \caption[]{}
         \label{fig:attribute_estimation_error}
     \end{subfigure}
     \hfill
     \begin{subfigure}[b]{0.49\textwidth}
         \centering
         \includegraphics[width=\textwidth]{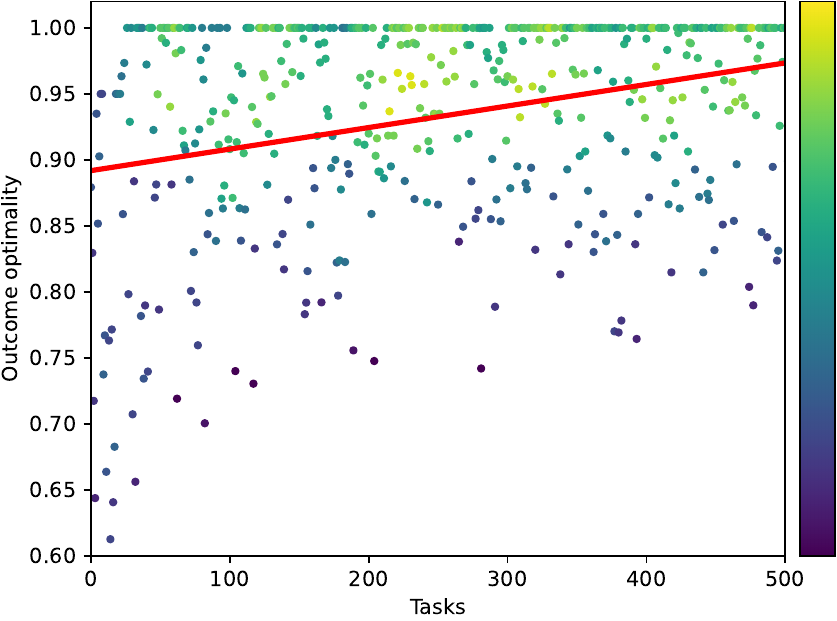}
         \caption[]{}
         \label{fig:task_outcome}
     \end{subfigure}
    \caption{In \ref{fig:human_requests} and \ref{fig:attribute_knowledge_evolution}, the horizontal axis is in logarithmic scale after time step 100. In \ref{fig:attribute_knowledge_evolution} and \ref{fig:attribute_estimation_error}, the vertical line at time step 1000 indicates when we started generating unbiased human feedback. In \ref{fig:task_outcome}, for visualization purposes, the color of each data point is proportional to the number of its neighbors, according to the colorbar. The red line is obtained via linear regression.
    }
    \label{fig:experiments}
\end{figure}

\textbf{Profiling performance.}
We start by showing the effectiveness of \LLMagent in estimating workers' latent attributes.
In our simulations the agent starts with an empty historical database and zero knowledge about workers' attributes to be estimated. The agent is only provided with the set $\Theta$ of possible attribute estimates, implicitly, using structured output generation: when generating profiling observations, the agent is often constrained to choose from a list of possible attributes, such as technical skills associated with specific job roles, and possible levels, according to the specified rating scale.
Due to this initialization, in the initial stages the available profiling data is often insufficient to evaluate some staffing criteria, making it necessary for the agent to ask workers questions and infer the missing information from their answers. Figure \ref{fig:human_requests} shows the number of questions asked to humans at each time step for the evaluation of criteria \ref{criterion:HSC} and \ref{criterion:TPC}, which address the proficiency in task-related technical skills and the enjoyment of tasks, respectively. As expected, the number of questions is highest in the early time steps and decreases as profiling data are collected through the analysis of human feedback.
Figures \ref{fig:attribute_knowledge_evolution} and \ref{fig:attribute_estimation_error} describe the evolution over time of the agent's knowledge about workers' attributes. In particular, Figure \ref{fig:attribute_knowledge_evolution} shows the number of correct, incorrect, and unknown estimates over time. Figure \ref{fig:attribute_estimation_error} shows the mean absolute estimation error (MAE) of workers' attributes over time, separately for each type of attribute. For unknown attributes, the estimation error is calculated with respect to a default estimate equal to the median level of the adopted rating scale.
Since the agent starts with no information about workers, most of the information collected during the initial time steps is completely new, resulting in a steep decrease in the number of unknown attributes. Later, the discovery rate of new attributes slows down and is mostly determined by the arrival of tasks to be staffed and by task assignment choices. In our simulations, some workers’ preferences are not discovered even after scheduling thousands of tasks, because tasks with certain characteristics have never been proposed to those workers.
Over time, the noise of individual observations is slowly smoothed out, resulting in progressively more accurate estimates. In the case of skill estimation, the biases that workers have towards each others result in steady state estimation errors. To verify the correct functioning of the agent in ideal scenarios, we generate unbiased but still noisy human feedback starting from time step 1000, which is highlighted in the plots by a vertical dotted line. As expected, removing biases causes the estimation error to decrease dramatically in subsequent time steps.

\textbf{Staffing performance.}
We now evaluate the staffing capabilities of \LLMagent and the benefits of worker profiling on the quality of the generated schedules. To do so, we need a way to measure the value associated with different staffing choices. In our simulations, the fitness of a task assignment is defined as the optimality of the resulting task outcome. Task outcomes are scalar values computed by an oracle by evaluating criteria \ref{criterion:SSD}, \ref{criterion:TMC} and \ref{criterion:HSC} using the true attributes of the workers assigned to the task. This outcome simulation model finds motivation in the literature, since maximizing these criteria was shown to promote better job and team performance \cite{ivancevich1990organizational, tekleab2016re}. The optimality of a task assignment is obtained by dividing the resulting task outcome by the maximum achievable task outcome at the time of scheduling, considering all the possible scheduling alternatives.
Figure \ref{fig:task_outcome} shows the optimality score of the task assignments generated by \LLMagent during the first 500 time steps. The scatter plot suggests a clear improvement over time in the staffing performance of the agent, which is confirmed by the linear regression estimator in red. As expected, the optimality of task assignments grows as workers' attributes estimates become more accurate, sharing a common trend.

%% file: sections/related_work.tex
\section{Related Work}
\label{sec:related_work}

Our work can be linked to multiple research areas: LLM agents, user profiling, task scheduling and assignment, and team formation. We now review the recent research efforts in these areas, illustrating the gaps in the literature that our work fills and our unique and unprecedented contributions.

Task scheduling and personnel assignment have been directly and indirectly investigated within very different research fields, each with their perspectives, tools and research questions. As a result, there is a wealth of relevant results scattered across the literature, some of which can become complementary pieces of a more complex system. Building on this idea, in this paper we also show that LLM agents have great potential to combine insights from different disciplines into a unified framework.

\subsection{LLM Agents and LLM-Based Approaches}

\textbf{LLM agents}.
Autonomous agents can be defined as situated systems capable of sensing the surrounding environment and accordingly taking intentional actions to pursue their objectives, without the need of external help. Based on this definition, LLM agents are autonomous agents whose brain is an LLM. They can be represented as complex interconnected systems composed of multiple modules that extend the capabilities of a central LLM. Typically, four main modules are present \cite{wang2024survey}: a profile module assigning an identity to the LLM, a memory module allowing access to stored data and reflection on past behaviors, a planning module responsible for decision-making and strategy definition, and an action module to execute the plan in the environment.

\textbf{LLMs for staffing and hiring}.
To the best of our knowledge, the only existing work in which LLMs are applied to staffing problems is \cite{mikhridinova2024using}, in which an LLM is prompted to analyze the compatibility of two predefined teams for a specific project. However, their study is very limited as it only considers a single staffing choice, and even the author themselves state that the resulting decision support is unusable. LLMs have been more widely applied to another related problem arising in work environments, namely recruitment. For example, the automated resume screening proposed in \cite{gan2024application} uses an LLM to summarize resumes, rank them and choose the best candidate. The job recommendation framework in \cite{du2024enhancing} uses an LLM to improve the quality of resumes by leveraging workers' interactions with job advertisements.

\textbf{LLMs for user profiling.}
In our work we seek to learn implicit features of the workers, such as their true skill levels and preferences towards tasks and teammates. This can be seen as a type of user profiling, which aims to derive a structured representation of the interests, characteristics, behaviors and preferences of a user \cite{purificato2024user}. Looking at the LLM-based user profiling literature, \cite{wang2025know} uses an LLM to infer implicit user profiles from human-machine conversations and generate more personalized and realistic dialogues.
PALR \cite{yang2023palr} asks an LLM to summarize the interaction history of users to extract user preferences. Similarly, the recommendation system in \cite{shu2024rah} uses LLMs to understand which generic features associated with items are liked or disliked by specific users based on their feedback. A review of LLM-based user modeling and profiling is available in \cite{tan2023user}. We do not directly mention approaches to user modeling and profiling that do not leverage LLMs as they are unrelated from out work, and refer the reader to the survey \cite{purificato2024user}.

\textbf{LLMs for recommendation systems.}
We conclude by mentioning a few applications of LLMs to recommendation systems, a topic related to both staffing and profiling. RecMind \cite{wang2023recmind} is an LLM agent for general zero-shot recommendation that exploits a self-inspiring planning algorithm, while InteRecAgent \cite{huang2023recommender} is an LLM agent for conversational recommendation that employs recommendation models as tools. The LLM assistant in \cite{shu2024rah} utilizes user preferences to filter the recommendations generated by a standard recommendation system, and infers user preferences from user feedback.

% Besides LLMs, other approaches to user modeling and user profiling have been proposed / lines of research in the field of machine learning take human preferences as the subject to be learned. RLHF learns a reward function that represent preferences.

\subsection{Task Scheduling and Team Formation}

\textbf{Optimization.}
The optimization community formulated and addressed a variety of problems about personnel assignment and project scheduling. Several works address multi-skill project or multi-project scheduling, in which the objective is to assign workers with multiple skills and proficiency levels to various project activities, focusing on different aspects: \cite{akbar2022multi} considers the personality traits of workers, \cite{li2020multi} considers skill development and cooperation effectiveness, \cite{chen2017multi} considers the economic cost and time associated with scheduling and skill learning-forgetting, \cite{chen2012ant} considers precedence constraints between tasks and combines ant colony optimization with an event-based scheduler.
MNA \cite{li2015metanetwork} uses a three-dimensional metanetwork representing relationships between personnel, knowledge and tasks to address a project task assignment problem.
In \cite{lagesse2006game} a modified Gale-Shapely courtship algorithm is proposed to assign tasks to employees based on skills and available time of each employee and preferences of both managers and employees.
Using fuzzy optimization, \cite{shen2003multi} assigns tasks to employees based on their skill set, social relationship score and work history.\\
In general, optimizations approaches lead to rigorous problem formulations and algorithmic efficiency, but lack flexibility and have limited descriptive power. In fact, the above works can be seen as model-based solutions: they make multiple assumptions, use variables and coefficients whose values are difficult to estimate, and require rigid descriptions of tasks, workers, and optimization criteria thus solving only very specific variants of the problem.

\textbf{Machine learning.}
The task scheduling machine learning literature focuses almost exclusively on machines, drones and robotic agents, and only a few works use machine learning to address team formation. Among them, GCO-DQN \cite{lv2024team} combines reinforcement learning and graph neural networks to determine how to select the subset of workers to be removed from an organizational network and to be assigned to a different project so to minimize the impact on the network. RELEXT \cite{chang2022learning} applies reinforcement learning to a subgraph extraction problem to find the group of experts satisfying skill coverage with minimized communication cost. The team formation approach of \cite{hamidi2020learning} uses a variational Bayes neural architecture trained on past team compositions, with dense vector representations of skill and expert subsets.

\textbf{Psychological and social sciences.}
Task assignment and team composition have also been analyzed from a humanistic perspective, formalizing intuitive concepts and deriving useful principles that can improve these processes. The general consensus is that organizational decisions should take into account personality, attitude and individual characteristics of workers to increase job performance and job satisfaction levels \cite{ivancevich1990organizational}. Several studies analyze the usefulness of standard psychometric instruments and personality tests for work purposes. Among them, \cite{barrick1991big} investigates the correlation between the Big Five personality traits \cite{mccrae1992introduction} (openness, conscientiousness, extraversion, agreeableness, neuroticism) and three job performance criteria. Another prominent aspect is the relationship between team diversity and team performance: team heterogeneity favors brainstorming and creative innovation, but may have detrimental effects on team cohesion \cite{tekleab2016re, garcia2017diversity}.

%% file: sections/conclusions.tex
\section{Conclusions and Future Work}
In this work, we proposed a novel formalization of the joint staffing and profiling problem and introduced \LLMagent, an LLM agent that solves this problem in an efficient and fully automated way. Differently from existing staffing solutions, \LLMagent allows expressing optimization objectives using natural language, accepts human-friendly task descriptions rather than variable-based formulations, removes the need for custom cost functions, and can process unstructured data directly. By analyzing human feedback and other data obtained as a by-product of staffing, the agent continuously estimates latent features of workers, such as their skills and personal preferences, realizing life-long worker profiling and improving the staffing performance over time. \LLMagent directly interacts with workers, allowing them to discuss assignments and provide real-time feedback, making the process intuitive, understandable and manageable by humans.
The internal design of \LLMagent ensures interpretability: our staffing procedure provides a clear separation of roles between scheduler, LLM reasoning and scripts, and the use of annotations in profiling allows tracing the estimation process of each attribute. Moreover, the LLM is prompted to briefly justify all its choices, promoting explainability. By providing tailored task assignments that consider workers’ skills and expertise, that are learned and updated over time, \LLMagent delivers a robust, transparent, and human-centric framework for automated personnel management.

\LLMagent is an innovative concept of LLM agent with great potential and scope for extensions and improvements in various aspects. Some interesting possibilities for future work include exploiting the code generation capabilities of LLMs, augmenting task descriptions using the built-in knowledge of LLMs and Retrieval-Augmented Generation, and learning the preferences of supervisors. An improved version of \LLMagent could integrate more advanced estimation techniques and consider additional profiling objectives. To give an example, in \cite{pandey2020work} work overload was the most frequent reason behind stress and had a significant adverse association with the performance of workers, which motivates estimating workers' stress levels to dynamically adjust their work capacity accordingly.

\subsection*{Acknowledgments}
The author was partially funded by Maschio Gaspardo S.p.A. and by Fondazione Ing. Aldo Gini. The author would like to thank Prof. Luca Schenato from the University of Padova for useful discussions and suggestions to improve the current paper.

%% file: sections/scheduler.tex
\section{A Possible Scheduler} \label{appendix:scheduler}
\begin{figure}[htbp]
  \centering
  \includegraphics[width=0.85\linewidth]{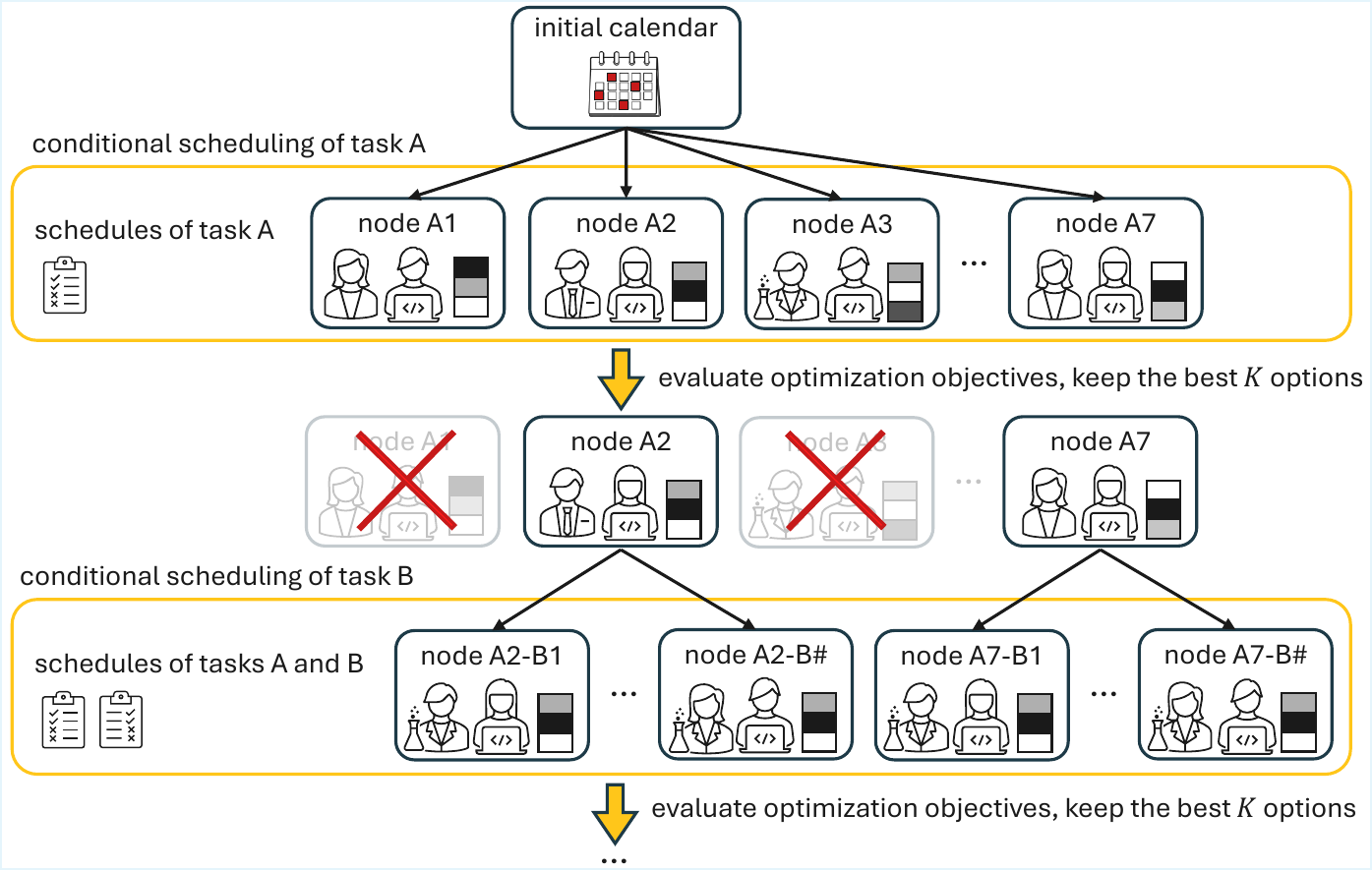}
  \caption{Visual representation of the multi-path conditional scheduling heuristic implemented in Algorithm \ref{alg:scheduler}.}
  \label{fig:scheduler}
\end{figure}

In this section we describe a possible algorithm to solve \eqref{eq:MIP} based on a multi-path conditional planning heuristic, that iteratively grows and prunes a tree to simultaneously explore multiple possibilities. Because it is based on a heuristic, this algorithm trades off optimality to gain computational efficiency, providing a good approximate solution in a short time. 
Importantly, this algorithm does not support constraint \eqref{eq:constraint_deadline} and is therefore not suitable for tasks that have a strict deadline. Note that \LLMagent treats the scheduler as a black-box, so the proposed algorithm can be seamlessly replaced with another of choice based on the specific use case. The main idea behind the algorithm is visually represented in Figure \ref{fig:scheduler}.

\textbf{One task at a time.}
The algorithm does not consider all the tasks simultaneously, as this would entail solving a combinatorial problem. Rather, tasks are processed one at a time following a given order, decomposing the problem and decreasing the number of possible assignment alternatives. For example, in our experiments we consider tasks associated with a priority level representing their relative importance and urgency, and order them as follows: we first sort tasks according to their priority level, and then sort tasks with the same priority according to their arrival time. This is motivated by the fact that a task with higher priority that arrived sooner should be given better treatment, and subsequent tasks should get the best remaining options according to their order. The order can be chosen based on the specific application, for example to facilitate the evaluation of precedence constraints. Despite tasks are considered sequentially, the algorithm is still able to jointly optimize multiple task assignments, as described in the following.

\boldcode{feasibleTSOs}$\mathbb{(\cdot)}$.
The function \code{feasibleTSOs}$(\cdot)$ returns a set of task scheduling options (TSOs) for a given task $t$, where each TSO is a pair $\bigl( \text{team} \subseteq W_k, [\alpha_t, \beta_t) \in \mathcal{I}_k \bigr)$ that specifies a subset of workers and a time interval. In particular, the function first identifies all the teams of workers that satisfy the requirements of the task, and then looks for the earliest time interval in which each team is available to perform the task, possibly within a finite planning horizon. This is done by overlapping the calendars of the members of a team, and sliding a time window until either a sufficiently long time interval is found or the planning horizon is reached. Keeping a finite planning horizon helps in preventing frequent rescheduling, which is always undesirable and can lead to sub-optimal staffing and resource management. \code{feasibleTSOs}$(\cdot)$ also checks for the satisfaction of constraints, and filters out unfeasible combinations of workers. Optionally, the function can receive an additional calendar to be considered on top of the others, that can be used to add further constraints.

\textbf{Multi-path planning.}
Recalling Definition \ref{def:feasible_schedule}, a schedule is a map assigning each task to a TSO. Intuitively, if multiple good TSOs of a task can be found, it is desirable to choose the one that suits subsequent tasks better. Implementing this principle, Algorithm \ref{alg:scheduler} uses a tree to keep track of multiple good scheduling alternatives, where each node stores a possible schedule. Accordingly, in this section the terms node and schedule are used interchangeably. The algorithm starts by initializing the tree with a root containing the initial calendar, e.g. the previous schedule $s_{k-1}$. It then invokes \code{feasibleTSOs}$(\cdot)$ on the first task in the list, obtaining the corresponding set of TSOs. If no option can be found, the algorithm simply goes to the next task. Each TSO results in a different children node of the root, containing the schedule in the root plus the new pair (new task, TSO). This originates independent paths that will be explored in parallel.
Consistent with the general workflow of the staffing module, the schedules stored in these newly created leaf nodes are evaluated using the dedicated functions $u_i(\cdot)$ and the aggregation function $V(\cdot)$. Instead of simply selecting the best schedule, the algorithm keeps up to $K$ schedules with highest aggregated score, where $K$ is a user-defined parameter, and prunes the other leaf nodes. When scheduling the next task in the list, \code{feasibleTSOs}$(\cdot)$ is called separately for each leaf node, each time passing the schedule stored in the node as the additional calendar argument taken by the function. This results in multiple sets of TSOs for the second task, each conditioned on a specific TSO of the first task. Each overall schedule, i.e. first plus second task, undergoes the same process as above, namely creation of associated children nodes, evaluation of the optimization criteria for each leaf node and selection of up to $K$ best schedules. Therefore, the scheduling of a task can be influenced by subsequent tasks, reducing the occurrence of greedy and short-sighted task assignments. This process continues iteratively until all tasks have been processed, growing hypothetical paths where each step is conditioned on previous choices and resulting in up to $K$ schedules. The algorithm is described using a tree for convenience of presentation, but the procedure can be implemented in more computationally and memory efficient ways.

\algorithmstyle{ruled}
\begin{algorithm}
\DontPrintSemicolon
\SetNoFillComment
\SetArgSty{textnormal}
\SetCommentSty{mycommentfont}
\SetKwInOut{Input}{input}
\Input{$T_k$, set of pending tasks sorted in decreasing order of importance.\newline
$W_k$, workers' state.\newline
$s_{k-1}$, current schedule.\newline
$K$, maximum number of schedules to be generated.\newline
Planning horizon for \code{feasibleTSOs()} (optional).
}
\vspace{4pt}
Initialize a tree with a root containing the current schedule.\par
\For{each task $t \in T_k$}{
    \tcp{Each node of the tree is a schedule}
    \For{each leaf $l$ in the tree}{
        \tcp{Generate scheduling alternatives for $t$ conditioned on schedule $l$}
        TSOs$(t) \leftarrow$  \code{feasibleTSOs}$(t, W_k \ | \  l)$\par
        \For{each TSO in TSOs$(t)$}{
            $l.\code{addChildren}(l \ \cup$ TSO$)$
        }
    }
    Evaluate the optimization objectives for each leaf and keep only the best $K$ schedules.
}
\KwRet{the $K$ leaves of the tree.}
\caption{schedule$(T_k,\ W_k,\ s_{k-1},\ K)$}
\label{alg:scheduler}
\end{algorithm}

\algorithmstyle{ruled}
\begin{algorithm}
\DontPrintSemicolon
\SetNoFillComment
%\LinesNumbered
\SetArgSty{textnormal}
\SetCommentSty{mycommentfont}
\SetKwInOut{Input}{input}
\SetKw{GoTo}{go to line}
\SetKw{Continue}{continue}
\Input{$T_k$, set of pending tasks.\newline
$W_k$, workers' state.\newline
$s_{k-1}$, current schedule.\newline
$m$, maximum number of rescheduling attempts for each task.
}
\vspace{4pt}
$s \leftarrow s_{k-1}$\par
Remove problematic tasks from $s$ and add them to $T_k$.\par
\While{$T_k$ is not empty}{
    \label{algline:while}    
    \vspace{-3mm}\noindent\parbox{0mm}{
    $$t \leftarrow \argmax_{t' \in T_k} \; t'\code{.priority}$$
    } \tcp*{Task with highest priority} \par \vspace{-2mm}
    TSOs$(t) \leftarrow$ \code{feasibleTSOs}$(t, W_k \ | \  s)$\par 
    \vspace{-3mm}\noindent\parbox{0mm}{
    $$\alpha_t^{min} \leftarrow \min_{ \bigl( \text{team},\ [\alpha_t, \beta_t) \bigr) \in \text{TSOs}(t)} \, \alpha$$
    } \tcp*{Current earliest start time of $t$} \par \vspace{-2mm}
    $j \leftarrow 0$ \tcp*{Rescheduling attempt index} \par
    $h \leftarrow k$ \tcp*{Window start timestep index} \par
    $i \leftarrow 1$ \tcp*{Number of tasks in each combination} \par
    $C_{old} \leftarrow \varnothing$\par
    \While{\code{True}}{
        \If{$P_h$ is undefined}{
            \vspace{-3mm}\noindent\parbox{0mm}{
            \begin{align*}
                P_h \leftarrow \bigl\{ & t' \ | \ t' \in s,\ \alpha_{t'} \geq k,\ [h,\ h+t\code{.duration}) \cap [ \alpha_{t'}, \beta_{t'}) \neq \varnothing, \\
                & t'\code{.priority} < t\code{.priority},\ t'\code{.requirements} \cap t\code{.requirements} \neq \varnothing \bigr\}
            \end{align*}
            }\par \vspace{-2mm}
        }
        $C \leftarrow \bigl\{ c \ | \ c \text{ is a combination of } i \text{ tasks} \in P_h \bigr\}$\par
        \For{each combination $c \in (C \setminus C_{old}).\code{sort}()$}{
            $j \leftarrow j + 1$\par
            \uIf{$t$ can be scheduled in place of $c$}{
                Remove tasks in $c$ from $s$ and add them to $T_k$.
            }
            \ElseIf(\tcp*[f]{Up to $m$ rescheduling attempts}){$j < m$}{
                \Continue
                }
            $s \leftarrow \code{schedule}(t, W_k, s, 1)$ \tcp*{No rescheduling} \par
            \KwRet{\code{reschedule}$(T_k \setminus \{t\}, W_k, s)$}
        }
        $h \leftarrow h + 1$\par
        $C_{old} \leftarrow C_{old} \cup C$\par
        \If(\tcp*[f]{Consider larger task combinations}){$h \geq \alpha_t^{min}$}{
            $h \leftarrow k$\par
            $C_{old} \leftarrow \varnothing$\par
            $i \leftarrow i + 1$ \par
            \If{$i \geq \max_{h \in [k, \alpha_t^{min}-1]} \; |P_h|$}{
                $s \leftarrow \code{schedule}(t, W_k, s, 1)$ \tcp*{No rescheduling} \par
                \KwRet{\code{reschedule}$(T_k \setminus \{t\}, W_k, s)$}
                }
            }
        }
    }
\KwRet{$s$}
\caption{reschedule$(T_k,\ W_k,\ s_{k-1},\ m)$}
\label{alg:rescheduling}
\end{algorithm}

\textbf{Rescheduling.}
In some cases, such as the arrival of a new high-priority task or changes in worker availability, it may be needed to modify an existing schedule. Obtaining the best schedule given the new conditions would require rescheduling all tasks from scratch, which in most cases is unacceptable as humans are involved. The priority is therefore to find the feasible schedule that respects the importance level of tasks and deviates as little as possible from the initial schedule. We now describe Algorithm \ref{alg:rescheduling}, a recursive procedure that approximately solves this problem by implementing two principles: \textit{(i)} cancel scheduled tasks if this allows a pending, higher priority task to be scheduled earlier, \textit{(ii)} cancel as few tasks as possible. First, in case the initial schedule is infeasible, we cancel problematic tasks and put them back in the list of pending tasks. We then identify the pending task $t$ with highest priority, and compute its earliest feasible start time $\alpha_t^{min}$ given the current schedule. In the remainder of the algorithm we look for a subset of scheduled tasks that, if canceled, would allow $t$ to be scheduled before $\alpha_t^{min}$. If such a subset is found, the tasks it contains are removed from the schedule and added to the pending tasks, making room for task $t$. Otherwise, if a suitable task subset cannot be found within $m$ rescheduling attempts, where $m$ is a user-defined parameter, task $t$ is simply scheduled on top of the current schedule. In both cases, the updated schedule and list of pending tasks are given to a new instance of the algorithm in a recursive fashion. We now describe how to select the subset of tasks to be canceled according to the principles above. For each time step $h \in [k,\ \alpha_t^{min}-1]$, where $k$ is the current time step, let $P_h$ be the set of currently scheduled tasks that have not started yet, are scheduled in a time interval overlapping with $[h,\ h+t.\code{duration})$, have lower priority compared to $t$ and require resources needed also by $t$. To cancel as few tasks as possible, we first try to cancel single tasks, and then combinations of tasks of progressively larger size. For each $i \geq 1$, we consider canceling each of the different combinations made of $i$ tasks belonging to $P_h$, iterating over $h \in [k,\ \alpha_t^{min}-1]$. When reaching $h=\alpha_t^{min}$, we increase $i$ and restart from $h=k$, until a good combination is found, or the number of rescheduling attempts reaches $m$, or all combinations have been processed. To improve performance, it is recommended to order combinations based on the inconvenience caused by deleting tasks contained within them. For example, it may be convenient to first consider combinations containing low-priority tasks or tasks that are easy to reallocate.

%% file: sections/experiments_appendix.tex
\section{Supplementary Simulation Details}
\label{appendix:simulation_details}

\textbf{Simulating human feedback - supplement.}
In our experiments, we simulate four types of human feedback consistent with the model \eqref{eq:feedback_model}: self-evaluation feedback, task proposal feedback, performance reviews and peer feedback. Below, we describe how we synthetically generate each type of feedback.
\begin{itemize}
    \item Self-evaluation feedback. This is the response of workers to the questions posed by the staffing module when the available profiling data is not sufficient to evaluate certain optimization criteria. This type of feedback is simulated using a LLM, which is prompted to impersonate specific workers given a noisy and biased version of their true skill and preference levels.
    \item Task proposal feedback. This is the feedback accompanying the acceptance or rejection of the tasks proposed by the agent. The choice of accepting or rejecting tasks is taken algorithmically based on the preferences of workers to have better control over the simulation. More specifically, a task is rejected with a probability proportional to the worker's average level of aversion to various aspects of the task. Similarly as in the previous case, a LLM is prompted to impersonate target workers and briefly motivate the acceptance or rejection of the proposed task based on the assigned skills and preferences, that are noisy and biased.
    \item Performance review. This can be a report written by the project manager leading the team or a review left by the client that benefited of the task. To simulate it, we feed the task description and the attributes of the team members to a LLM and prompt it to first imagine a plausible outcome of the task, and then generate performance reviews in natural language format for each worker. Since the outcome of the task is expected to reflect the actual capabilities of workers, in this case we input a noisy but unbiased version of their attributes.
    \item Peer feedback. These are reviews of workers coming from their colleagues, collected for example at the end of team projects. To simulate peer feedback, we randomly choose a few template sentences from a set of options and fill them with competence levels, without resorting to a LLM. To realistically introduce subjectivity, these competence levels are a noisy and biased version of the true skill levels of the worker to be evaluated, where noise and bias differ for each reviewer according to \eqref{eq:feedback_model_b}. Peer feedback also contains a sentence expressing how much the reviewer enjoys working with the reviewed colleague, and this sentence is selected based on the true teammate preferences of the reviewer.
\end{itemize}

\textbf{Optimization criteria.}
In our staffing simulations we set the 9 optimization objectives described below. We consider a mix of quantitative objectives, evaluated using algorithms only, and qualitative objectives, expressed using natural language and evaluated with the aid of the LLM.
\begin{enumerate}
    \item Maximize the average priority level of the tasks in the schedule.
    \item Minimize the average waiting time of the tasks in the schedule. The waiting time of each task, defined as the difference between scheduled start time and arrival time of the task, is weighted by the priority level of the task.
    \item Maximize the number of tasks in the schedule.
    \item Minimize the economic cost of the schedule, computed by multiplying workers' wages by their working time.
    \item Minimize the variance between workers' active times, so that all workers are busy approximately the same amount of time.
    \item \label{criterion:SSD} In team tasks, maximize the diversity in soft skills among team members, measured using the cosine distance. More specifically, to obtain this metric we vectorize the soft skill levels of each worker, compute the cosine similarity between these vectors for each pair of workers in the team, rescale these values as $\psi \mapsto (1 - \psi)/2$ and compute the average.
    \item \label{criterion:TMC} In team tasks, maximize the compatibility between team members, computed using the estimated teammate preferences.
    \item \label{criterion:HSC} For each task, maximize the match between the task description and the hard skills of the workers assigned to it. This involves asking the LLM to compute a compatibility score for each task-worker pair based on the estimated hard skills of the workers. The LLM can answer with a special value to flag that the available information about the worker is insufficient. In such case, \LLMagent directly asks the worker about the unknown technical skill levels.
    \item \label{criterion:TPC} For each task, maximize the fit between the task description and the preferences of the workers assigned to it. Also in this case, we ask the LLM to compute a numerical score for each task-worker pair based on the estimated task preferences of workers. In case of insufficient information, \LLMagent assumes that workers do not have significant preferences or aversions for the task.
\end{enumerate}
Criterion \ref{criterion:SSD} promotes the selection of teams whose members have heterogeneous soft skill levels: this incentivizes balanced teams that cover the full spectrum of soft skills, while penalizing teams with incompatible redundancies, such as multiple workers with strong leadership. According to \cite{ivancevich1990organizational}, heterogeneous groups typically outperform homogeneous ones because they have a richer variety of knowledge and problem-solving strategies to draw from, but they may lack cohesion, thus motivating criterion \ref{criterion:TMC}.

%% file: main.bbl
\begin{thebibliography}{10}

\bibitem{ivancevich1990organizational}
J.~M. Ivancevich, M.~T. Matteson, and R.~Konopaske, {\em Organizational behavior and management}.
\newblock Bpi/Irwin New York, 1990.

\bibitem{isen1991positive}
A.~M. Isen and R.~A. Baron, ``Positive affect as a factor in organizational behavior,'' {\em Research in organizational behavior}, vol.~13, no.~1, pp.~1--53, 1991.

\bibitem{bahroun2024multi}
Z.~Bahroun, R.~As'~ad, M.~Tanash, and R.~Athamneh, ``The multi-skilled resource-constrained project scheduling problem: A systematic review and an exploration of future landscapes.,'' {\em Management Systems in Production Engineering}, vol.~32, no.~1, 2024.

\bibitem{mikhridinova2024using}
N.~Mikhridinova, F.~A. Yurdakul, and C.~Wolff, ``Using large language models for project staffing: Evaluation of gpt-based mapping of teams to projects,'' in {\em Proceedings of the 12th IPMA Research Conference “Project Management in the Age of Artificial Intelligence”}, pp.~1--19, IPMA, 2024.

\bibitem{tekleab2016re}
A.~G. Tekleab, A.~Karaca, N.~R. Quigley, and E.~W. Tsang, ``Re-examining the functional diversity--performance relationship: The roles of behavioral integration, team cohesion, and team learning,'' {\em Journal of business research}, vol.~69, no.~9, pp.~3500--3507, 2016.

\bibitem{wang2024survey}
L.~Wang, C.~Ma, X.~Feng, Z.~Zhang, H.~Yang, J.~Zhang, Z.~Chen, J.~Tang, X.~Chen, Y.~Lin, {\em et~al.}, ``A survey on large language model based autonomous agents,'' {\em Frontiers of Computer Science}, vol.~18, no.~6, p.~186345, 2024.

\bibitem{gan2024application}
C.~Gan, Q.~Zhang, and T.~Mori, ``Application of llm agents in recruitment: A novel framework for resume screening,'' {\em arXiv preprint arXiv:2401.08315}, 2024.

\bibitem{du2024enhancing}
Y.~Du, D.~Luo, R.~Yan, X.~Wang, H.~Liu, H.~Zhu, Y.~Song, and J.~Zhang, ``Enhancing job recommendation through llm-based generative adversarial networks,'' in {\em Proceedings of the AAAI Conference on Artificial Intelligence}, vol.~38, pp.~8363--8371, 2024.

\bibitem{purificato2024user}
E.~Purificato, L.~Boratto, and E.~W. De~Luca, ``User modeling and user profiling: A comprehensive survey,'' {\em arXiv preprint arXiv:2402.09660}, 2024.

\bibitem{wang2025know}
K.~Wang, X.~Li, S.~Yang, L.~Zhou, F.~Jiang, and H.~Li, ``Know you first and be you better: Modeling human-like user simulators via implicit profiles,'' {\em arXiv preprint arXiv:2502.18968}, 2025.

\bibitem{yang2023palr}
F.~Yang, Z.~Chen, Z.~Jiang, E.~Cho, X.~Huang, and Y.~Lu, ``Palr: Personalization aware llms for recommendation,'' {\em arXiv preprint arXiv:2305.07622}, 2023.

\bibitem{shu2024rah}
Y.~Shu, H.~Zhang, H.~Gu, P.~Zhang, T.~Lu, D.~Li, and N.~Gu, ``Rah! recsys--assistant--human: A human-centered recommendation framework with llm agents,'' {\em IEEE Transactions on Computational Social Systems}, 2024.

\bibitem{tan2023user}
Z.~Tan and M.~Jiang, ``User modeling in the era of large language models: Current research and future directions,'' {\em arXiv preprint arXiv:2312.11518}, 2023.

\bibitem{wang2023recmind}
Y.~Wang, Z.~Jiang, Z.~Chen, F.~Yang, Y.~Zhou, E.~Cho, X.~Fan, X.~Huang, Y.~Lu, and Y.~Yang, ``Recmind: Large language model powered agent for recommendation,'' {\em arXiv preprint arXiv:2308.14296}, 2023.

\bibitem{huang2023recommender}
X.~Huang, J.~Lian, Y.~Lei, J.~Yao, D.~Lian, and X.~Xie, ``Recommender ai agent: Integrating large language models for interactive recommendations,'' {\em arXiv preprint arXiv:2308.16505}, 2023.

\bibitem{akbar2022multi}
S.~Akbar, I.~Ahmad, R.~Khan, I.~O. Lopes, and R.~Ullah, ``Multi-skills resource constrained and personality traits based project scheduling,'' {\em IEEE Access}, vol.~10, pp.~131419--131429, 2022.

\bibitem{li2020multi}
Q.~Li, Q.~Sun, S.~Tao, and X.~Gao, ``Multi-skill project scheduling with skill evolution and cooperation effectiveness,'' {\em Engineering, Construction and Architectural Management}, vol.~27, no.~8, pp.~2023--2045, 2020.

\bibitem{chen2017multi}
R.~Chen, C.~Liang, D.~Gu, and J.~Y. Leung, ``A multi-objective model for multi-project scheduling and multi-skilled staff assignment for it product development considering competency evolution,'' {\em International Journal of Production Research}, vol.~55, no.~21, pp.~6207--6234, 2017.

\bibitem{chen2012ant}
W.-N. Chen and J.~Zhang, ``Ant colony optimization for software project scheduling and staffing with an event-based scheduler,'' {\em IEEE Transactions on Software Engineering}, vol.~39, no.~1, pp.~1--17, 2012.

\bibitem{li2015metanetwork}
Y.~Li, Y.~Lu, D.~Li, and L.~Ma, ``Metanetwork analysis for project task assignment,'' {\em Journal of Construction Engineering and Management}, vol.~141, no.~12, p.~04015044, 2015.

\bibitem{lagesse2006game}
B.~Lagesse, ``A game-theoretical model for task assignment in project management,'' in {\em 2006 IEEE International Conference on Management of Innovation and Technology}, vol.~2, pp.~678--680, IEEE, 2006.

\bibitem{shen2003multi}
M.~Shen, G.-H. Tzeng, and D.-R. Liu, ``Multi-criteria task assignment in workflow management systems,'' in {\em 36th Annual Hawaii International Conference on System Sciences, 2003. Proceedings of The}, pp.~9--pp, IEEE, 2003.

\bibitem{lv2024team}
B.~Lv, J.~Jiang, L.~Wu, and H.~Zhao, ``Team formation in large organizations: A deep reinforcement learning approach,'' {\em Decision Support Systems}, vol.~187, p.~114343, 2024.

\bibitem{chang2022learning}
C.-C. Chang, M.-Y. Chang, J.-Y. Jhang, L.-Y. Yeh, and C.-Y. Shen, ``Learning to extract expert teams in social networks,'' {\em IEEE Transactions on Computational Social Systems}, vol.~9, no.~5, pp.~1552--1562, 2022.

\bibitem{hamidi2020learning}
R.~Hamidi~Rad, H.~Fani, M.~Kargar, J.~Szlichta, and E.~Bagheri, ``Learning to form skill-based teams of experts,'' in {\em Proceedings of the 29th ACM international conference on information \& knowledge management}, pp.~2049--2052, 2020.

\bibitem{barrick1991big}
M.~R. Barrick and M.~K. Mount, ``The big five personality dimensions and job performance: a meta-analysis,'' {\em Personnel psychology}, vol.~44, no.~1, pp.~1--26, 1991.

\bibitem{mccrae1992introduction}
R.~R. McCrae and O.~P. John, ``An introduction to the five-factor model and its applications,'' {\em Journal of personality}, vol.~60, no.~2, pp.~175--215, 1992.

\bibitem{garcia2017diversity}
M.~Garcia~Martinez, F.~Zouaghi, and T.~Garcia~Marco, ``Diversity is strategy: the effect of r\&d team diversity on innovative performance,'' {\em R\&D Management}, vol.~47, no.~2, pp.~311--329, 2017.

\bibitem{pandey2020work}
D.~L. Pandey, ``Work stress and employee performance: an assessment of impact of work stress,'' {\em International Research Journal of Human Resource and Social Sciences}, vol.~7, no.~05, pp.~124--135, 2020.

\end{thebibliography}
